\documentclass[10pt,twocolumn,letterpaper]{article}

\usepackage[pagenumbers]{cvpr} % To force page numbers, e.g. for an arXiv version

\usepackage[dvipsnames]{xcolor}
\definecolor{darkred}{rgb}{0.7, 0.0, 0.0}

\usepackage{bbding}
\usepackage{tabularx}
\usepackage{overpic}
\definecolor{cvprblue}{rgb}{0.21,0.49,0.74}
\usepackage[pagebackref,breaklinks,colorlinks,allcolors=cvprblue]{hyperref}

\def\paperID{4815} % *** Enter the Paper ID here
\def\confName{CVPR}
\def\confYear{2025}

\title{Event-assisted 12-stop HDR Imaging of Dynamic Scene}

\author{Shi Guo\textsuperscript{1} \quad Zixuan Chen\textsuperscript{1,2} \quad Ziran Zhang\textsuperscript{1,2} \quad Yutian Chen\textsuperscript{1,3} \quad Gangwei Xu\textsuperscript{4} \quad  Tianfan Xue\textsuperscript{3,1}\\
\textsuperscript{1}Shanghai AI Laboratory; \textsuperscript{2}Zhejiang University; \\
\textsuperscript{3}The Chinese University of Hong Kong;
\textsuperscript{4}Huazhong University of Science and Technology
\\
\url{https://openimaginglab.github.io/Event-Assisted-12stops-HDR/}
}

\begin{document}

\twocolumn[{
\renewcommand\twocolumn[1][]{#1}
\maketitle
\begin{center}
    \centering
    \captionsetup{type=figure}
    \includegraphics[width=\linewidth]{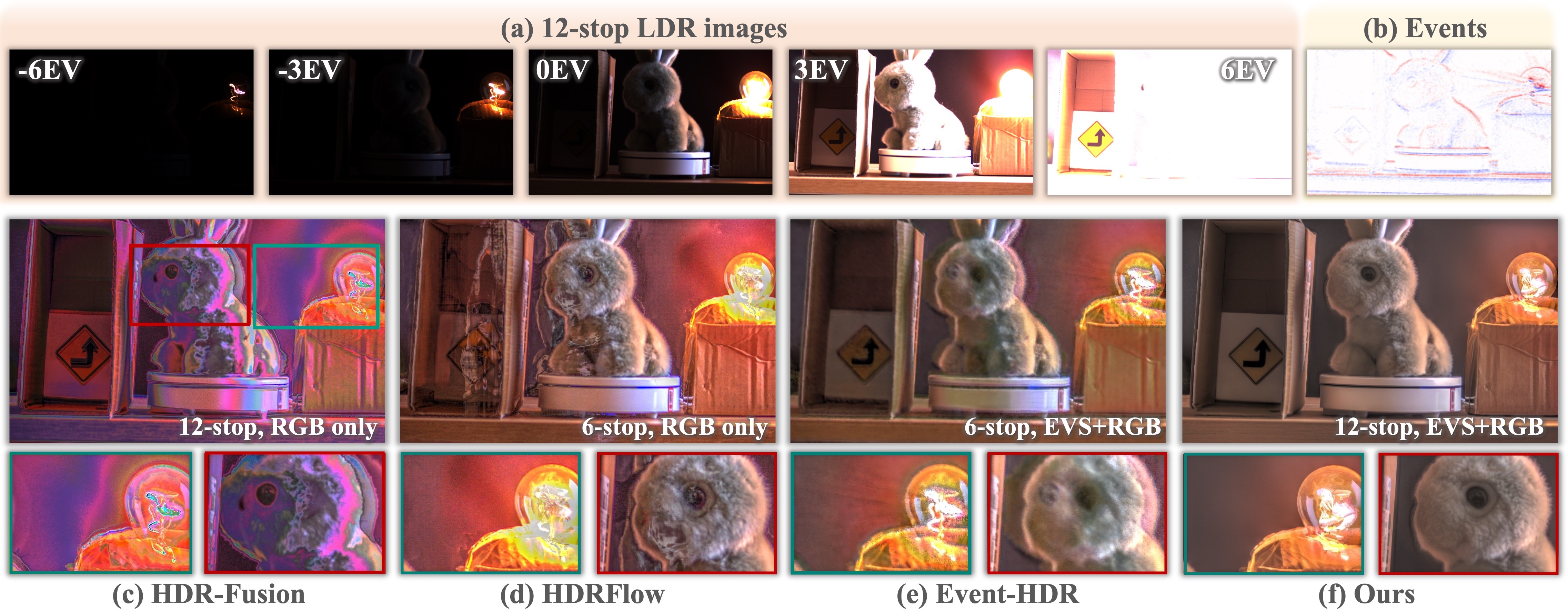}
    \vspace{-20pt}
    \captionof{figure}{Visual comparisons of different HDR imaging methods on real-captured data. We utilize \emph{Vibrant} tone mapping style in commercial HDR software (Photomatix) to better visualization.}
    \label{fig:teaser}
\end{center}%
}]

% \maketitle
\begin{abstract}
High dynamic range (HDR) imaging is a crucial task in computational photography, which captures details across diverse lighting conditions. Traditional HDR fusion methods face limitations in dynamic scenes with extreme exposure differences, as aligning low dynamic range (LDR) frames becomes challenging due to motion and brightness variation. In this work, we propose a novel 12-stop HDR imaging approach for dynamic scenes, leveraging a dual-camera system with an event camera and an RGB camera. The event camera provides temporally dense, high dynamic range signals that improve alignment between LDR frames with large exposure differences, reducing ghosting artifacts caused by motion. Also, a real-world finetuning strategy is proposed to increase the generalization of alignment module on real-world events. Additionally, we introduce a diffusion-based fusion module that incorporates image priors from pre-trained diffusion models to address artifacts in high-contrast regions and minimize errors from the alignment process. To support this work, we developed the ESHDR dataset, the first dataset for 12-stop HDR imaging with synchronized event signals, and validated our approach on both simulated and real-world data. Extensive experiments demonstrate that our method achieves state-of-the-art performance, successfully extending HDR imaging to 12 stops in dynamic scenes.

\end{abstract}

\section{Introduction}
\label{sec:intro}
High dynamic range (HDR) imaging is a fundamental task in computational photography, aiming to accurately capture the full range of visual details in both bright and dark regions of a scene. Since the typical dynamic range of an image sensor is much lower than real-world luminance ranges, which is often about 80 to 100dB~\cite{xiao2002high,pattanaik1998multiscale}, post-capture processing is often required to increase sensor dynamic range. The most common solution is to merge multiple low dynamic range (LDR) images captured at varying exposure levels, known as HDR fusion.

Although HDR fusion is widely used, it still faces challenges when applied to very high dynamic range scenes, like a night street with very bright light. These scenes typically require 7 to 12 exposure stops to fully recover the dynamic range~\cite{barakat2008minimal,gardner2017learning,hold2019deep}. Majority of HDR fusion algorithms may fail, as aligning images with very different exposures becomes challenging, considering potential variation between images due to either object motion or exposure difference. As a result, previous dynamic HDR methods are restricted to handling only 2–3 exposure images within a 6-stop range (\eg, -3EV to +3EV), leading to information loss in high-contrast scenes (see Fig.~\ref{fig:teaser} (d)). 
Given that, this raises an important question: \emph{Is it possible to extend HDR imaging to 12 stops in dynamic scenes?}

To solve the alignment challenges, we propose an event-RGB dual-camera HDR imaging system. The additional event camera~\cite{serrano2013128,brandli2014240,chen2024event} can perfectly solve the alignment problem in large exposure differences, because event cameras can capture temporally dense event streams for accurate alignment, and also have a much higher intrinsic dynamic range compared to normal RGB cameras (normally 60-80dB). As shown in Fig.~\ref{fig:teaser} (b), event signals provide consistent, fine-grained motion information across both bright (+6EV LDRs) and dark (-6EV LDRs) regions. Moreover, unlike previous approaches~\cite{messikommer2022multi,shaw2022hdr} that implicitly use event data to assist HDR fusion, our approach directly estimates motion between RGB frames from event streams. This design avoids ghosting artifacts as shown in~\cref{fig:teaser} (e), and allows us to handle much higher exposure differences (12 stops vs. 4-6 stops in previous approaches).

Another challenge is to robustly fuse LDR images with large exposure differences. Current learning-based HDR imaging methods~\cite{kalantari2019deep,liu2021adnet,chen2021hdr,liu2023joint,xu2024hdrflow} employ $\mu$-law tone mapping as a loss function. However, the $\mu$-law function alone is insufficient for effectively managing both bright and dark regions in 12-stop HDR. Moreover, in the presence of alignment error, the fusion module must also introduce artifacts, particularly at occlusion boundaries. To tackle these challenges, we propose a diffusion-based fusion module. It integrates pre-trained image priors to reduce error accumulation from the alignment module and mitigate artifacts in regions where texture details are highly compressed, as shown in \cref{fig:teaser} (f).

At last, there is no 12-stop HDR imaging with event signals. Therefore, we developed the ESHDR dataset, the first dataset for 12-stop HDR imaging with synchronized event signals, using a novel simulation pipeline. While previous approaches~\cite{kalantari2019deep,liu2021adnet,chen2021hdr,liu2023joint,xu2024hdrflow} simulate multiple exposure LDRs from LDR images, our simulation pipeline generates LDR images from -6 to +6EV using HDR scenes to ensure high-quality simulation. Furthermore, to validate our methods in real-world scenarios, we built a RGB-event dual-camera system capable of capturing synchronized Event and RGB sequences across a -6EV to 6EV range. Extensive experiments on both simulated and real-world datasets demonstrate that our framework achieves state-of-the-art performance across challenging scenes, successfully pushing HDR imaging to 12 stops.

Our key contributions are summarized as follows:
\begin{itemize}
    \item To the best of our knowledge, we present the first attempt at 12-stop HDR imaging for dynamic scenes.

    \item We propose an event-assisted diffusion-based HDR method that focuses on solving two major challenges in dynamic-scene 12-stop HDR imaging: image alignment and fusion for LDRs with extreme exposure differences.

    \item We develop the first dataset, ESHDR, for 12-stop HDR imaging with event signals, and build a dual-camera system to validate our approach on both simulated and real-world datasets, demonstrating the practical feasibility and effectiveness of our method.
\end{itemize}
\begin{figure*}[!t]
    \centering
    %\vspace{0.1cm}
    \begin{overpic}[width=1\textwidth]{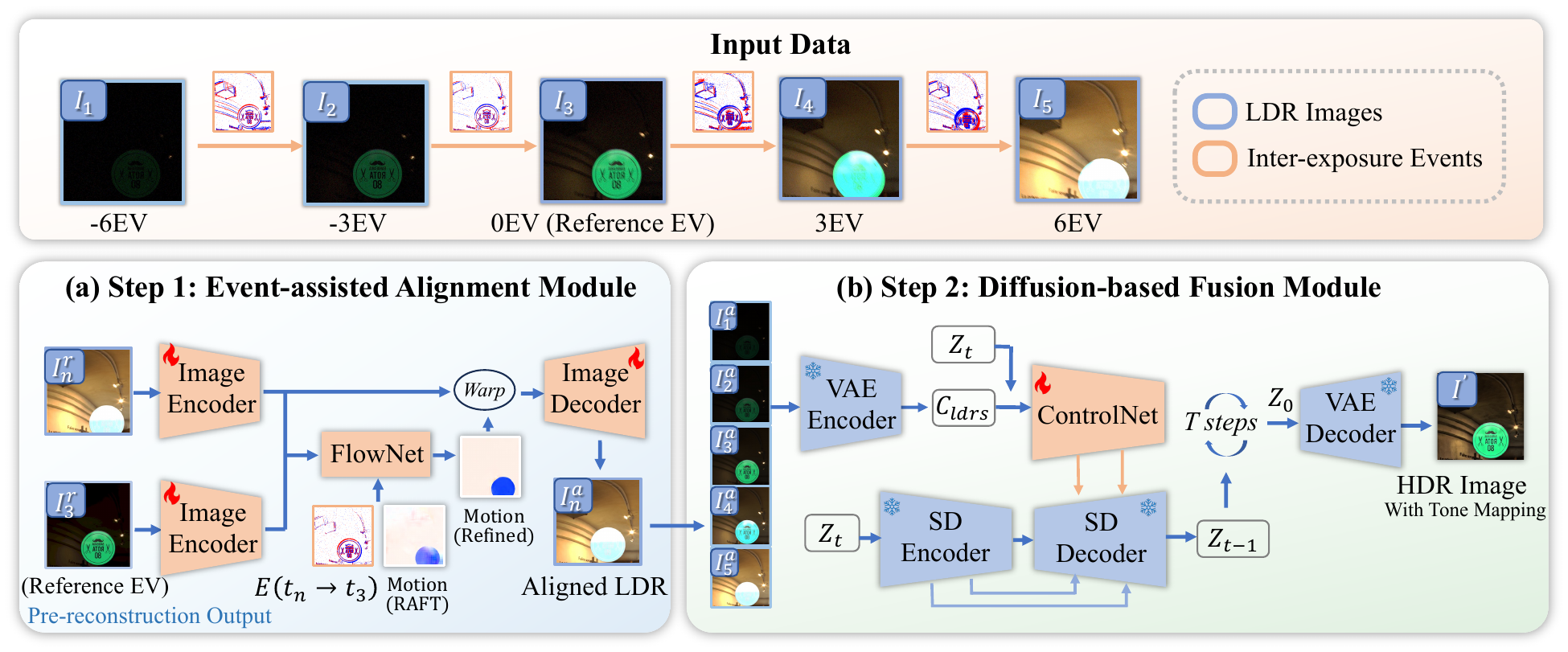}
    \end{overpic}
    \vspace{-10pt}
    \caption{Illustration of the proposed framework for 12-stop HDR Imaging in Dynamic Scenes. The framework consists of two main components: (a) an event-assisted alignment module and (b) a diffusion-based fusion module.}
    \label{fig:main-network}
\end{figure*}

\section{Related Work}
\label{sec:related-work}

\subsection{RGB-based HDR imaging}
To capture high dynamic range (HDR) images, the most common approach is to merge multiple low dynamic range (LDR) images taken at various exposure levels. This approach, however, presents challenges in dynamic scenes, as motion often leads to ghosting artifacts~\cite{hu2013hdr,oh2014robust}. 

Previous HDR methods for dynamic scenes~\cite{hu2013hdr,ma2017robust,oh2014robust,sen2012robust,mangiat2010high,wu2018deep,yan2019attention,yan2022dual,liu2022ghost,chen2022attention,liu2021adnet,chen2021hdr,liu2023joint,yan2023unified,chung2023lan,kalantari2019deep,xu2024hdrflow} have focused on mitigating these artifacts through advanced image alignment techniques. Specifically, \citet{kalantari2013patch} introduced a patch-based synthesis technique to improve temporal continuity in HDR images by enforcing similarity across adjacent frames. \cite{yan2019attention,yan2022dual} proposed a context-aware transformer to enhance model performance in complex dynamic scenes. Additionally, methods using deformable convolutions~\cite{liu2021adnet,chen2021hdr} and more sophisticated models like the pyramid cross-attention module~\cite{liu2023joint} and content alignment model~\cite{yan2023unified} with ghost attention and patch aggregation have been developed. \cite{chung2023lan} explored frame alignment through a luminance-based attention score, while \cite{kalantari2019deep,xu2024hdrflow} used motion flow estimation from LDRs to facilitate alignment through warping.

Despite these advancements, RGB-based HDR methods still face limitations in aligning LDRs with large exposure differences and complex motion due to the lack of motion information between frames. Some multi-exposure fusion methods~\cite{mertens2007exposure,ma2015multi,kou2017multi,ma2019deep} can achieve 12-stop HDR, but they are only effective in static scenes, which significantly limits their applicability. In this paper, we employ a high-dynamic-range event camera to aid alignment and, for the first time, tackle the challenging and valuable task of 12-stop HDR imaging in dynamic scenes.

\subsection{Event-based HDR imaging}
Event camera captures pixel-level intensity changes with exceptional temporal precision, offering a dynamic range up to 140 dB~\cite{serrano2013128, brandli2014240}. This high temporal resolution makes event cameras ideal to assist frame alignment in HDR imaging in dynamic scenes.

Early works explored using event streams to directly generate HDR videos or to assist in reconstructing HDR images from single LDR frames. Generative adversarial networks (GANs)~\cite{wang2019event} and recurrent networks~\cite{rebecq2019high} have been employed to generate HDR videos directly. \citet{yang2023learning} incorporated event signals to guide the reconstruction of HDR images from LDR frames, enhancing image quality in dynamic environments. Additionally, \citet{xiaopeng2024hdr} introduced a self-supervised framework that leverages event data for motion compensation, significantly reducing motion blur in HDR reconstruction. However, due to the lack of spatial information in event data and the inherent noise and degradations within event signals, the results remain suboptimal. Recent approaches have extended event-based HDR methods by integrating multi-exposure LDR images alongside event data. \citet{messikommer2022multi} utilized deformable convolutions to align multi-exposure LDR frames, while \citet{shaw2022hdr} proposed an event-to-image feature distillation module for better fusion of event data with RGB images. However, these methods implicitly use event data for 4-6 stop HDR alignment, struggling with large motion resulting in ghosting artifacts.

Our work addresses these challenges by proposing an explicit event-assisted alignment module. By fine-tuning the alignment module on real-world interpolation datasets, our approach achieves improved alignment performance in real scenes. Furthermore, we introduce a diffusion-based fusion module specifically designed to handle extreme exposure differences, pushing HDR imaging in dynamic scenes to a 12-stop range for the first time.

\section{Method}

\subsection{Overall framework}
This study tackles the complex problem of 12-stop high dynamic range (HDR) reconstruction in dynamic scenes from a sequence of $n$ low dynamic range (LDR) images, $ \{I_1, I_2, \ldots, I_N\}$, each captured at distinct timestamps $\{t_1, t_2, \ldots, t_N\}$, with corresponding exposure times $\{\tau_1, \tau_2, \ldots, \tau_N\}$. Specifically, we choose 5 LDR images for fusion with exposures $\{$-6EV, -3EV, 0EV, 3EV, 6EV$\}$.
Additionally, this sequence is supplemented by a continuous stream of events, $\{E_i\}_{t_1}^{t_N}$, providing HDR temporal detail. Our objective is to achieve exposure fusion and synthesize a 12-stop HDR image, selectively integrating fine details across exposures. The obtained HDR image aligns with the reference frame $I_{\text{ref}}$, identified at timestamp $t_{\text{ref}}$.

In contrast to previous methods in exposure fusion~\cite{mertens2007exposure,ma2015multi,kou2017multi,ma2019deep} and dynamic HDR imaging~\cite{kalantari2019deep,liu2021adnet,chen2021hdr,liu2023joint,xu2024hdrflow}, we introduce a novel approach capable of performing 12-stop HDR imaging in dynamic scenes. To address the inherent challenges—specifically, the frame alignment and fusion of LDR frames with substantial exposure variation—we propose leveraging HDR event signals to facilitate explicit frame alignment in \cref{sec:event-assisted-reconstruction}. Additionally, we incorporate a diffusion-based fusion model to generate the 12-stop HDR image in \cref{sec:diff_fusion}. By utilizing the image prior learned from a pre-trained diffusion model, our fusion process achieves enhanced visualization in regions where textures are compressed within a narrow dynamic range, while also mitigating alignment-related error accumulation. The overall pipeline is summarized in \cref{fig:main-network}.

\subsection{Event-assisted alignment}
\label{sec:event-assisted-reconstruction}
An event camera captures changes in scene illumination as an asynchronous stream of events. Each event at a position $ \mathbf{u} = (x, y) $ is triggered when the relative change in intensity exceeds a contrast threshold $ c $:
\begin{equation}
    E_t = \begin{cases}
        1, & \text{if } \log(I_{t}) - \log(I_{t-\Delta t}) \geq c, \\ 
        -1, & \text{if } \log(I_{t}) - \log(I_{t-\Delta t}) \leq -c,\\        
        0, & \text{otherwise},
        \end{cases}       
\label{eq:normal_define_evs}
\end{equation}
where $E_t$ represents the polarity of the event, $I_{t}$ denotes the image intensity at time $t$, and $\Delta t$ is the interval since the last event at position $\mathbf{u}$. 

By recording illumination changes in the $\log$ domain, the event signal provides not only microsecond-level temporal resolution but also a high dynamic range reaching up to 140 dB. This enables the event signal to capture motion information between LDR images and to preserve details in both bright and dark regions of RGB images, spanning a range of -6EV to 6EV. We leverage this event signal to support frame alignment.

\noindent \textbf{Pre-reconstruction.}
At very high dynamic scene, the input LDR images often contain noise and motion blur, which will hurt the alignment and fusion quality. To mitigate this issue, we first train a UNet model~\cite{ronneberger2015u} that recovers a clean and sharp LDR image $I^{r}_n$ from the input $I_n$. For better reconstruction, we utilize the event signal $E_{t_n \rightarrow t_n + \tau_n}$ during the exposure $\tau_n$ of the LDR RGB image. Since our training process employs a synthesized data pipeline, the ground-truth LDR image $\hat{I}_n$ is available for supervision. The loss function used for training is:
$\mathcal{L}_{\text{pre}} = \sqrt{\Vert \hat{I}_n - I^{r}_n \Vert^2 + \epsilon^2}$,
where $\sqrt{\Vert \hat{x} - x \Vert^2 + \epsilon^2}$ represents the Charbonnier penalty function with $\epsilon$ set to 0.001.

\begin{figure*}[t!]
    \centering
\includegraphics[width=\textwidth]{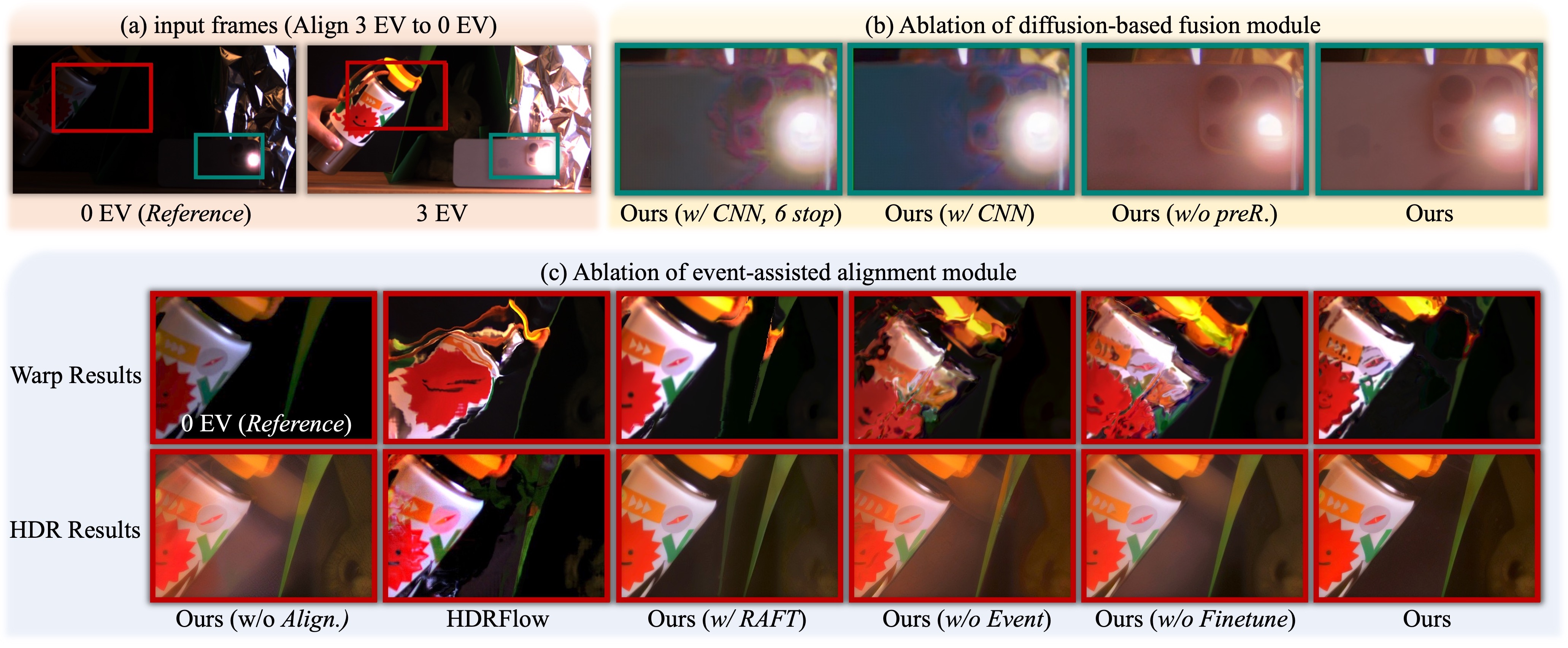}
    \vspace{-15pt}
    \caption{Ablation study on fusion and alignment modules by comparison of different configurations.}
    \label{fig:ablation}
\end{figure*}

%\subsubsection{Inter-exposure Alignment}
\vspace{2pt}
\noindent \textbf{Event-assisted alignment module.} In the next step, we align multiple LDR images. Instead of the implicit alignment used in the previous event-based HDR~\cite{serrano2013128,brandli2014240}, we design an explicit event-assisted alignment module, as shown in~\cref{fig:main-network} (a). This is because with implicit alignment, previous video reconstruction literature~\cite{mildenhall2018burst} shows that the network may converge to a local minimum of a simple merging without alignment, because alignment is a more difficult task. In dynamic HDR imaging, simple merging may cause ghosting artifacts when LDRs have challenging exposure differences (see \cref{fig:teaser} (e)). Therefore, we use an explicit alignment, which aligns each $I^{r}_n$ with the reference LDR $I^{r}_{\text{ref}}$, using events between them.
% Using events between LDRs, denoted as $E_{t_n \rightarrow t_{\text{ref}}}$, to align each $I^{r}_n$ with the reference LDR $I^{r}_{\text{ref}}$ is important step for dynamic HDR methods. As discussed in the previous video reconstruction literature~\cite{mildenhall2018burst}, with an implicit alignment, the network may converges to a local minimum of a simple merging without alignment, because alignment is a more difficult task. In dynamic HDR imaging, simple merging may cause the ghosting artifact when LDRs have challenging exposure difference (see Fig.~\ref{fig:teaser} (e)). Thus we design a explicit event-assisted alignment module, which is shown in Fig.~\ref{fig:main-network} (a). 

% Refer to the success of alignment module in video reconstruction methods~\cite{wang2019edvr,chan2021understanding,guo2022differentiable}, our alignment module also choose to perform alignment in feature space. 
We perform the alignment in the feature space, following the convertion of video reconstruction~\cite{wang2019edvr,chan2021understanding,guo2022differentiable}.
The encoder features of the two LDR images are firstly extracted from the input:
%\begin{equation}
%f_n = \text{Image-Encoder}(I^r_n), \quad f_{\text{ref}} = \text{Image-Encoder}(A(I^r_{\text{ref}})),
%\end{equation}
\begin{equation}
\small
f_n, f_{\text{ref}} = \text{Image-Encoder}(I^r_n), \text{Image-Encoder}(A(I^r_{\text{ref}})),
\end{equation}
where $\text{Image-Encoder}(\cdot)$ represents the image encoder, which consists of two residual blocks, and $A(\cdot)$ is the exposure alignment which aligns the exposure of the reference LDR to $I^r_n$. Then the flow is estimated as:
\begin{equation}
v_c = \text{FlowNet}(f_n, f_{\text{ref}}, v_{\text{raft}}, E_{t_n \rightarrow t_{\text{ref}}}),
\end{equation}
where $v_{\text{raft}}$ is the initial flow estimated using the pre-trained RGB flow estimation model, RAFT~\cite{teed2020raft}. The final aligned frame is obtained by:
\begin{equation}
I^a_n = \text{Image-Decoder}(\mathcal{W}(f_n, v_c)),
\end{equation}
where $\mathcal{W}(x, v)$ performs backward warping using the optical flow $v$. $\text{Image-Decoder}(\cdot)$ denotes the decoder that converts the image features back into the image space. 
% To further enhance the accuracy of flow estimation, $f_n$ and $f_{\text{ref}}$ are divided into four groups, and the flow is estimated separately for each group.

\vspace{2pt}
\noindent \textbf{Real-world fine-tuning.}
As discussed in \cite{kim2024frequency,duan2024led,zhang2024sim}, even with state-of-the-art event simulators, v2e~\cite{hu2021v2e} or ESIM~\cite{rebecq2018esim}, a distribution gap remains between simulated and real-world events. Previous methods~\cite{tulyakov2021time,kim2024frequency,ma2025timelens} have collected real-world training pairs to improve generalization in real-world scenes. However, obtaining 12-stop HDR training pairs in dynamic scenes with event data is very challenging. Therefore, we choose to fine-tune the event-assisted alignment module using real-world interpolation datasets, which are easier to acquire and widely available. 

We utilize real-world datasets from \cite{ma2025timelens,cho2024tta}, which were collected using the same event camera as in our setup, the Prophesee EVK4-HD. The two input RGB frames, $I_1$ and $I_2$, are randomly converted to LDR images with varying exposure times between -6EV and 6EV, denoted as  $I_1^{\text{ev1}}$ and $I_2^{\text{ev2}}$. The goal is to align the output of the alignment module with $I_2^{\text{ev1}}$. Compared with only use synthesis data for training (\cref{fig:ablation}, Ours (w/o Finetune)), our method achieves improved results in real-world scenes and reduces ghosting artifacts by fine-tuning the alignment module.

\vspace{2pt}
\subsection{Diffusion-based fusion} 
\label{sec:diff_fusion}
In 12-stop HDR fusion, many details are compressed into a narrow value range. This compression causes even minor discrepancies in the fusion output to produce significant artifacts after tone mapping. Moreover, errors and occlusions in the alignment can further impact the fusion process. As shown in \cref{fig:ablation}, these challenges cause simply using a convolutional network (Our method (w/ CNN)) introduces noticeable artifacts in the fusion results. To address these issues, we introduce the image prior of pre-trained diffusion model (StableDiffusion v2.1)~\cite{Rombach_2022_CVPR} in the fusion process and propose a diffusion-based fusion module. Details can refer to
\cref{fig:main-network} (b).

Since the latent stable diffusion module is pre-trained on LDR images, we generate the latent representation $z_0 \in \mathbb{R}^{h \times w \times c}$ by applying variational autoencoder (VAE)~\cite{kingma2013auto} on the tone-mapped HDR image $\Gamma(I') \in \mathbb{R}^{H \times W \times 3}$, where $h$, $w$, and $c$ represent the dimensions of the latent feature and $\Gamma(\cdot)$ is the $\mu$-law tone mapping function. In the forward diffusion process, Gaussian noise is progressively added to $z_0$, resulting in a noisy latent representation $z_t$ at each timestep $t \in \{0, 1, \dots, T\}$. In the reverse diffusion process, this noise is gradually removed, starting from standard Gaussian noise. At timestep $t$, the aligned LDR images $I_n^a$ are used as the conditioning input, represented by $z_a = \text{VAE-Decoder}(\{I_n^a\}_{n=1}^N)$. The network $\epsilon_{\theta}$ then predicts the added noise with the following objective:
\begin{equation}
    \mathcal{L}_{\text{df}} = \mathbb{E}_{z_0, z_a, t, \epsilon \sim \mathcal{N}(0, 1)} \left[ \left\| \epsilon - \epsilon_{\theta}(z_t, z_a, t) \right\|_2^2 \right],
\end{equation}
where $\mathcal{L}_{\text{df}}$ is the diffusion fusion loss. To control HDR image generation using aligned LDRs while preserving pretrained diffusion model’s prior, we follow \cite{zhang2023adding} and use a trainable copy of the UNet encoder layer as the ControlNet to process the aligned LDRs. The output $z_t$ is then fed into the frozen UNet of the diffusion model. As shown in \cref{fig:ablation} (b), by using pre-trained latent stable diffusion model to provide an image prior, our method achieves artifact-free HDR results, with better fusion in the highlight regions.

%\begin{table*}[th]
%\centering
%\begin{tabular}{l|l|l|l|l}
%        & RGB Noise                 & RGB Blur                  & Single EV Range                            & Event                     \\ \hline
%HDRFlow & $\checkmark$ &                           & {[}0ev, -3ev{]}, {[}-2ev, 0ev, 2ev{]}      &                           \\
%HDRev   &                           &                           & Single EV                                  & $\checkmark$ \\ \hline
%Ours    & $\checkmark$ & $\checkmark$ & {[}-6ev, -4ev, -2ev, 0ev, 2ev, 4ev, 6ev{]} & $\checkmark$
%\\ \hline
%\end{tabular}
%\end{table*}
\section{RGB-Event HDR dataset}
\begin{figure}[t!]
\setlength{\belowcaptionskip}{-0.2cm}
    \centering
\includegraphics[width=0.47\textwidth]{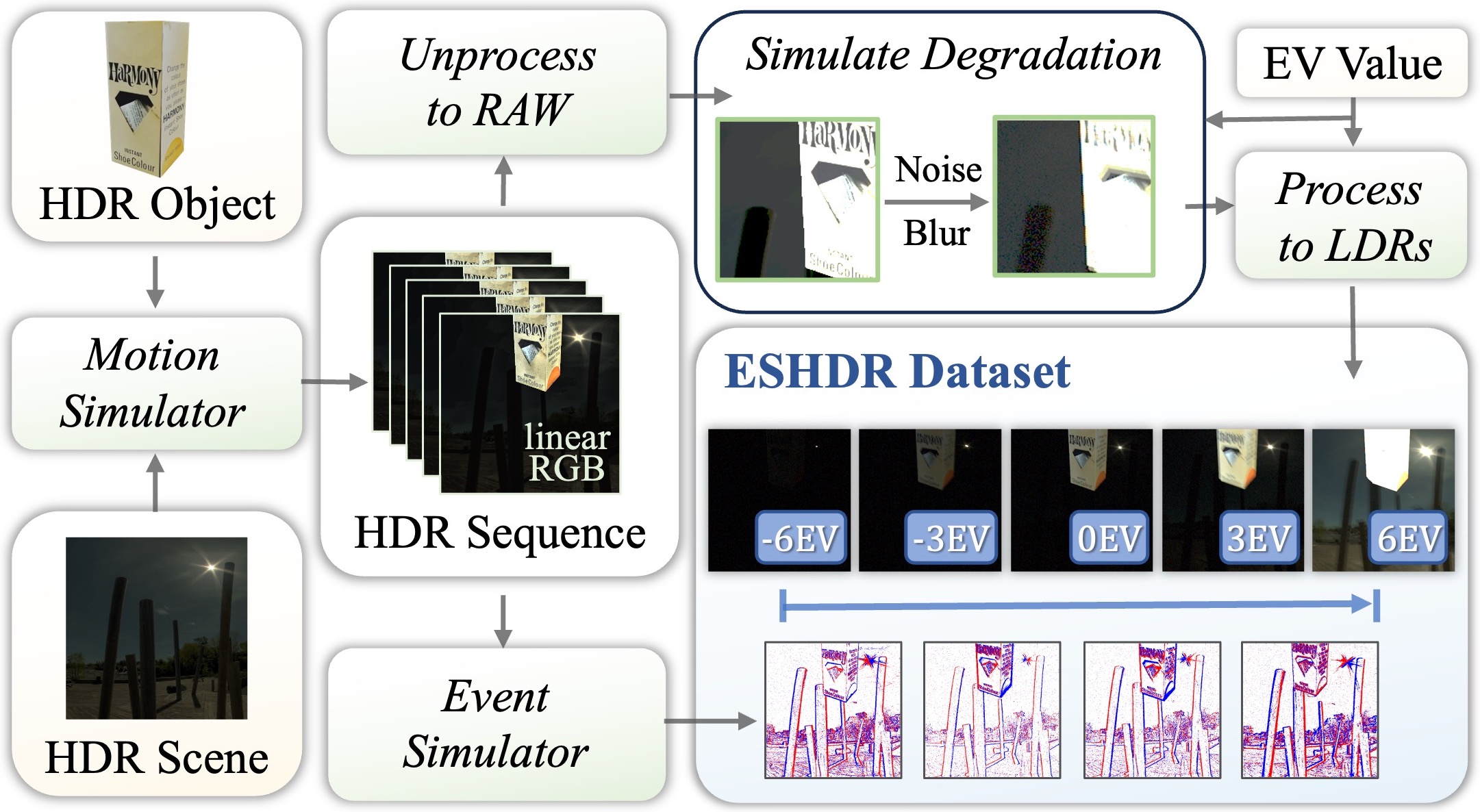}
    \vspace{-0pt}
    \caption{Simulation pipeline of 12-stop HDR imaging with event.}
    \label{fig:data_simulation}
\end{figure}

Due to the lack of dataset for 12-stop HDR imaging in dynamic scene with event signal, we developed a simulation pipeline and obtain the ESHDR dataset for training and evaluation. As illustrated in \cref{fig:data_simulation}, we first simulate high-frame-rate HDR sequences to generate events and LDRs with varying EV levels. The HDR sequence is created by blending a real HDR foreground object and background scene, both with random motion. Unlike previous methods~\cite{kalantari2019deep,liu2021adnet,chen2021hdr,liu2023joint,xu2024hdrflow}, which simulate LDRs from well-exposed LDR, our approach uses HDR sequences to ensure that the simulated LDRs from -6EV to 6EV contain as much complementary information as possible. 

The LDRs from -6EV to 6EV are then simulated from the HDR sequences. To simulate real-world noise and blur, these HDR images are first converted to RAW space, where Poisson-Gaussian noise~\cite{brooks2019unprocessing,guo2019toward} is added. Next, HDR frames are blended according to their exposure times to simulate motion blur. Finally, based on the EV values, brightness clipping and quantization are applied to produce LDRs at different EV levels. Additionally, event signals are generated from the high-frame-rate HDR sequences in linear color space, using the event simulator v2e~\cite{hu2021v2e}, to obtain high dynamic range event data.

Specifically, HDR scenes are sourced from the HDR Indoor~\cite{gardner2017learning} and Outdoor~\cite{hold2019deep} datasets, while HDR objects are obtained from the Objaverse dataset~\cite{deitke2023objaverse}. These datasets are divided into non-overlapping training and testing sets.

\begin{table*}[!t]
\centering
\renewcommand{\arraystretch}{1.2}
\caption{Comparison of different methods on HDR image quality metrics. Higher values indicate better performance for metrics marked with ($\uparrow$), while lower values are preferable for metrics marked with ($\downarrow$). Note that $\mu$-PSNR, $\mu$-SSIM, LPIPS, DISTS, and MANIQA are evaluated in the $\mu$-law tone-mapped domain.}
\label{tab:hdr_comparison}
\begin{tabular}{l|cccccc|c}
\hline
\textbf{Method} & \textbf{PSNR} & \textbf{SSIM} & \textbf{$\mu$-PSNR} & \textbf{$\mu$-SSIM} & \textbf{LPIPS} ($\downarrow$) & \textbf{DISTS} ($\downarrow$) & \textbf{MANIQA} ($\uparrow$) \\\hline
HDRFlow (3-stop)       & 31.34 & 0.9653 & 27.57 & 0.9131 & 0.1938 & 0.1729 & 0.2519 \\
HDRFlow (6-stop)      & 32.14 & 0.9689 & 28.70 & 0.9152 & 0.1655 & 0.1599 & 0.2581 \\
HDR-Transformer (6-stop) & 32.38 & 0.9700 & 29.06 & 0.9153 & 0.1710 & 0.1607 & 0.2470 \\
HDR-Fusion (12-stop) & 27.38 &0.9493 &23.14 &0.8275 &0.3383 &0.2388 &0.2369\\
\hline
HDRV* (single LDR, Event)      & 30.81 & 0.9746 & 31.87 & 0.9533 & 0.1269 & 0.1443 & 0.2996 \\
Event-HDR* (6-stop, Event) & 33.85 & 0.9857 & 34.43 & 0.9523 & 0.1190 & 0.1425 & 0.3050 \\
\hline
Ours* (12-stop, Event)    & \textbf{38.25} & \textbf{0.9876} & \textbf{35.66} & \textbf{0.9541} & \textbf{0.1091} & \textbf{0.1187} & \textbf{0.3168} \\\hline
\end{tabular}
\end{table*}

\section{Experiments}

\subsection{Experiment details}
\label{sec:exp_details}
In our experiment, we train our network on the training set of ESHDR. The model is implemented using PyTorch. During training, we use the Adam optimizer~\cite{Kingma2014AdamAM} with learning rate of $1 \times 10^{-4}$. We first train the event-assisted alignment module to obtain aligned LDR images for different exposures. Then, we train the diffusion-based fusion module. To maintain the color fidelity of the output HDR images in ESHDR evaluation, we further fine-tune one layer of the VAE decoder for several steps, which we refer to as Ours*. More details are provided in the supplementary material.

\subsection{Baseline methods}
To evaluate our model, we first compare it with state-of-the-art RGB dynamic scene HDR imaging methods, namely HDRFlow~\cite{xu2024hdrflow} and HDR-Transformer~\cite{liu2022ghost}. Since HDRFlow and HDR-Transformer can only fuse LDR images ranging from -3EV to 3EV, we also compare our approach with a state-of-the-art HDR fusion method, HDR-Fusion~\cite{debevec2023recovering}, which can process 12-stop HDR imaging. To avoid color shifts between different RGB HDR imaging methods, we align the colors of all results to the ground truth HDR using a 3D bilateral grid~\cite{wang2024bilateral}. For event-assisted HDR imaging methods, due to different experimental settings and the unavailability of released code, we re-trained HDRV~\cite{yang2023learning} and Event-HDR~\cite{messikommer2022multi} on our training set, which we denote as HDRV* and Event-HDR*. 
%It is worth noting that the original versions of HDRV and Event-HDR do not account for the blurring and noise effects present in LDR images.

\begin{figure*}[t!]
    \centering
\includegraphics[width=\textwidth]{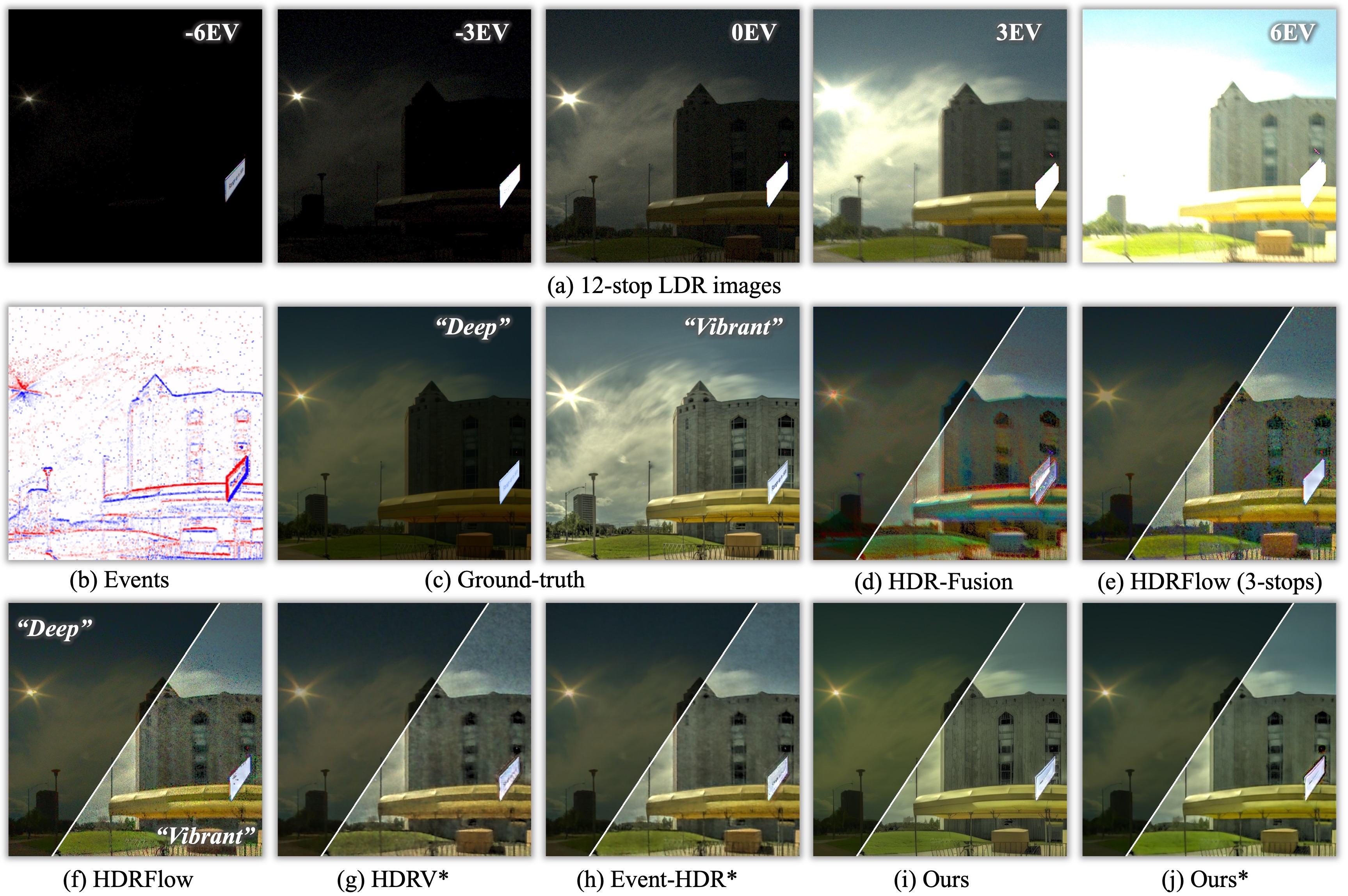}
    \vspace{-20pt}
    \caption{Visual comparisons of different HDR imaging methods. Since visualizing both the darkest and brightest areas in a 12-stop HDR image is a challenging task, we utilize two tone mapping styles, namely \emph{Deep} and \emph{Vibrant}, using commercial HDR software (Photomatix) to better highlight details in the bright and dark regions for comparison.}
    \label{fig:exp_syn_1}
\end{figure*}

\begin{figure*}[t!]
\setlength{\belowcaptionskip}{-0.0cm}
    \centering
\includegraphics[width=\textwidth]{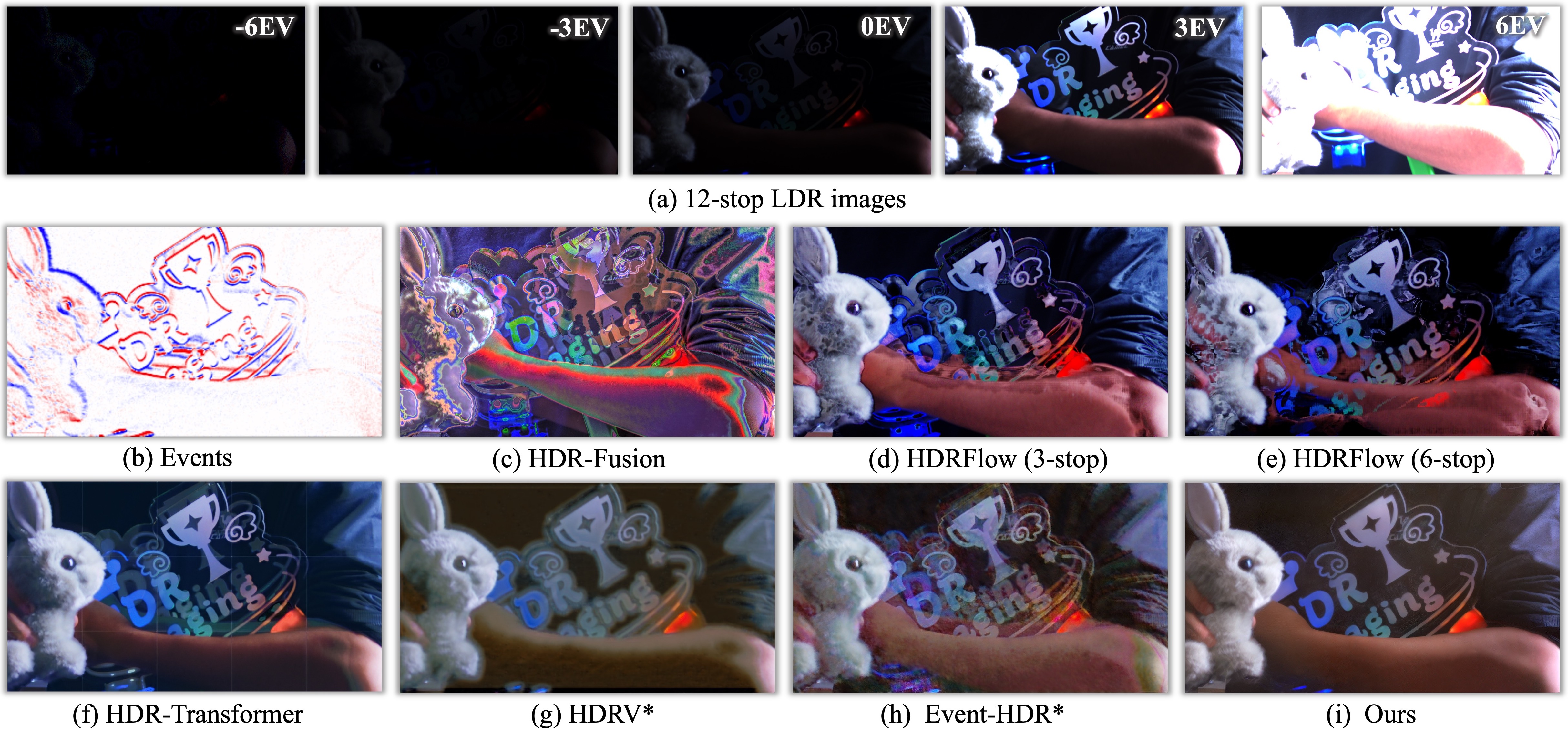}
    \vspace{-20pt}
    \caption{Visual comparisons of different HDR imaging methods on real-captured data. We utilize tone mapping in commercial HDR software (Photomatix) to better visualization.}
    \label{fig:exp_real_1}
\end{figure*}

\subsection{Experiments on ESHDR dataset}
In \cref{tab:hdr_comparison} and \cref{fig:exp_syn_1}, we present the quantitative and visual comparison results on the test set of the ESHDR dataset. It can be observed that our method achieves the best performance in both fidelity metrics (PSNR and SSIM) and perceptual quality metrics (LPIPS~\cite{zhang2018perceptual}, DISTS~\cite{ding2020iqa}, and non-reference MANIQA~\cite{yang2022maniqa}). 

In \cref{fig:exp_syn_1}, we also observe that using more LDR images for fusion allows HDRFlow (with -3EV to 3EV exposures) to achieve better results in bright regions compared to HDRFlow using 0EV to 3EV exposures. Although HDR-Fusion employs 12-stop LDR images for fusion, this method is limited to static scenes, resulting in ghosting artifacts in dynamic scenes. For HDRV*, even though it utilizes high dynamic range event signals, relying on a single LDR image is insufficient to achieve visually pleasing results. For Event-HDR*, despite leveraging event signals and alternately exposed LDR images, it performs better than previous methods in HDR recovery. However, in scenes with extremely high dynamic ranges, it remains noticeable artifacts (see the \emph{Vibrant} tone mapping style) in regions where details are compressed into narrow regions.

With the natural image priors embedded in our diffusion-based approach, our HDR reconstruction effectively reduces artifacts in both bright and dark regions. Furthermore, thanks to the fusion of LDR images ranging from -6EV to 6EV, our method delivers superior results in extremely high dynamic range regions (\eg, the Sun in \cref{fig:exp_syn_1}), achieving minimal overexposure in bright areas.

\subsection{Real-world experiments}

To validate our method in real-world scenes, we built a dual-camera system consisting of a Prophesee EVK4-HD ($1280 \times 720$) event camera and a RGB camera ($2048 \times 1536$). Our system ensures that both cameras are geometrically calibrated and temporally synchronized. By adjusting the exposure time of the RGB camera, we are able to capture LDR images from -6EV to 6EV along with the corresponding event signals. We constructed a real-world dataset consisting of 16 scenes, on which our algorithm achieved best performance on the widely-used non-reference metrics MUSIQ~\cite{ke2021musiq}, MANIQA~\cite{yang2022maniqa}, and HyperIQA~\cite{su2020blindly} (see \cref{tab:hdr_real_comparison}). Visual results from real-world scenes are shown in \cref{fig:exp_real_1}. It can be observed that this scene contains a high dynamic range and exhibits large motion, where the event camera successfully records useful signals in both the brightest and darkest regions.

Previous RGB-based HDR methods, such as HDRFlow and HDR-Transformer, exhibit noticeable artifacts in dynamic scenes despite performing alignment, due to the lack of motion information between LDR frames. Even using event signals, methods like Event-HDR*, which are trained only on simulated data, still display ghosting artifacts when tested in real-world scenes. For single-frame HDR methods with event assistance, such as HDRV*, the absence of complementary information and the misalignment between event and RGB edges due to motion prevent these methods from achieving satisfactory results. By fine-tuning the alignment module with real-world interpolation data and employing diffusion-based fusion, our method not only preserves more visually pleasing details in bright regions but also effectively reduces ghosting in areas with large motion.

\begin{table}[!t]
\setlength{\belowcaptionskip}{-0.1cm}
\centering
\renewcommand{\arraystretch}{1.2}
\caption{Quantitative comparison on real-world HDR imaging using non-reference quality metrics. Higher values indicate better performance for all metrics (MUSIQ, MANIQA, and HyperIQA). HDR-Trans. denotes HDR-Transformer.}
\label{tab:hdr_real_comparison}
\begin{tabular}{l|ccc}
\hline
\textbf{Method} & MUSID  & MANIQA & HyperIQA \\
\hline
HDRFlow (3-stop) & 39.46 & 0.2394 & 0.3266 \\
HDRFlow (6-stop) & 39.10 & 0.2001 & 0.3081 \\
HDR-Trans. & 39.72 & 0.2338 & 0.3351 \\
% HDR-Fusion & 45.87 & 0.2982 & 0.3697 \\
HDR-Fusion & 47.32 & 0.2981 & 0.3691 \\
\hline
HDRV* & 31.17 & 0.1639 & 0.2567 \\
HDR-Event* & 26.83 & 0.1527 & 0.2972 \\
\hline
Ours & \textbf{50.53} & \textbf{0.3011} & \textbf{0.4179} \\
\hline
\end{tabular}
\end{table}

\section{Discussion}
\label{sec:disscusion}
We discuss the roles of the primary modules in our method separately below: event-assisted alignment and diffusion-based fusion modules. Ablation results are in Fig.~\ref{fig:ablation}.
%The ablation results are shown in Fig.~\ref{fig:ablation}.

\noindent \textbf{Event-assisted alignment.} 
In Fig.~\ref{fig:ablation} (c), we compare our alignment module with various alignment methods. When the alignment module is completely removed (Ours w/o Align.), noticeable ghosting artifacts remain in the fusion results, even with the prior information from the diffusion model. Replacing our alignment module with the state-of-the-art optical flow estimation method RAFT shows good alignment when the brightness variation is small. However, it introduces significant misalignment in regions with large brightness changes as RAFT is not trained on them, resulting in structure errors in the fusion output.

To illustrate how event signals handle large motion effectively, we compared our method with HDRFlow, an RGB-based HDR method, and our variant without event signals (Ours w/o Event). In regions with large motion, both approaches fail to achieve accurate alignment due to the lack of motion information, leading to visible artifacts in the fusion results. By incorporating event signals (Ours w/o Finetune), our method achieves more accurate alignment, but a gap remains between simulated and real-world data, which limits its effectiveness in real scenes. After fine-tuning on a real-world RGB-Event interpolation dataset, our final method achieves reliable alignment and produces ghost-free fusion results.

\noindent \textbf{Diffusion-based fusion.} To evaluate the diffusion-based fusion module, we train two CNN-based fusion modules: Ours (w/ CNN, 6 stop) and Ours (w/ CNN). As shown in Fig.~\ref{fig:ablation} (b), CNN-based fusion struggles to eliminate alignment errors from the first stage. Additionally, for regions requiring an extremely high dynamic range, even though Ours (w/ CNN) uses more EV levels compared to Ours (w/ CNN, 6 stop), it still fails to produce visually pleasing results, particularly in scenes with intense lighting. By using the image priors from a pre-trained diffusion model, our method can achieve visually pleasing 12-stop HDR images. 

We also evaluate the effectiveness of the pre-reconstruction module by excluding pre-reconstruction (Ours w/o preR.). With the pre-reconstruction module, our method preserves more detailed textures, since this module reduces the noise in the input frames.

\noindent \textbf{Limitations.} 
Our dual-camera system, which uses a beam splitter, makes the acquisition setup less accessible to general users. We hope that advancements in hardware, such as hybrid event cameras~\cite{guo20233,kodama20231}, will enable 12-stop HDR imaging to become more widely applicable. Additionally, we did not specifically address motion blur in high EV levels, which we plan to consider in future work.
\section{Conclusion}
In conclusion, we have presented a novel approach for 12-stop HDR imaging in dynamic scenes, addressing the key challenges of alignment and fusion for LDR images with extreme exposure differences. By leveraging event signals from a dual-camera system, our method enhances alignment accuracy across high-contrast scenes, effectively mitigating ghosting artifacts caused by large motion. Also, by using real-world RGB-Event interpolation dataset to fine-tuning, our event-assisted alignment module can generalize well to real-world scenes. Furthermore, our diffusion-based fusion module incorporates pre-trained image priors to reduce artifacts in compressed detail regions and minimize error accumulation from alignment. To support our research, we developed the ESHDR dataset, the first dataset for 12-stop HDR imaging with synchronized event signals, and validated our approach using both simulated and real-world data. Experimental results demonstrate that our method achieves state-of-the-art performance in complex HDR scenarios, pushing the boundaries of dynamic HDR imaging to capture the full 12-stop range.

{
    \small
    \bibliographystyle{ieeenat_fullname}
    \bibliography{main}

\begin{thebibliography}{66}
\providecommand{\natexlab}[1]{#1}
\providecommand{\url}[1]{\texttt{#1}}
\expandafter\ifx\csname urlstyle\endcsname\relax
  \providecommand{\doi}[1]{doi: #1}\else
  \providecommand{\doi}{doi: \begingroup \urlstyle{rm}\Url}\fi

\bibitem[Barakat et~al.(2008)Barakat, Hone, and Darcie]{barakat2008minimal}
Neil Barakat, A~Nicholas Hone, and Thomas~E Darcie.
\newblock Minimal-bracketing sets for high-dynamic-range image capture.
\newblock \emph{IEEE Transactions on Image Processing}, 17\penalty0 (10):\penalty0 1864--1875, 2008.

\bibitem[Brandli et~al.(2014)Brandli, Berner, Yang, Liu, and Delbruck]{brandli2014240}
Christian Brandli, Raphael Berner, Minhao Yang, Shih-Chii Liu, and Tobi Delbruck.
\newblock A 240×180 130 db 3 us latency global shutter spatiotemporal vision sensor.
\newblock \emph{IEEE Journal of Solid-State Circuits}, 49\penalty0 (10):\penalty0 2333--2341, 2014.

\bibitem[Brooks et~al.(2019)Brooks, Mildenhall, Xue, Chen, Sharlet, and Barron]{brooks2019unprocessing}
Tim Brooks, Ben Mildenhall, Tianfan Xue, Jiawen Chen, Dillon Sharlet, and Jonathan~T Barron.
\newblock Unprocessing images for learned raw denoising.
\newblock In \emph{Proceedings of the IEEE/CVF conference on computer vision and pattern recognition}, pages 11036--11045, 2019.

\bibitem[Chan et~al.(2021)Chan, Wang, Yu, Dong, and Loy]{chan2021understanding}
Kelvin~CK Chan, Xintao Wang, Ke Yu, Chao Dong, and Chen~Change Loy.
\newblock Understanding deformable alignment in video super-resolution.
\newblock In \emph{Proceedings of the AAAI conference on artificial intelligence}, pages 973--981, 2021.

\bibitem[Chen et~al.(2021)Chen, Chen, Guo, Liang, Wong, and Zhang]{chen2021hdr}
Guanying Chen, Chaofeng Chen, Shi Guo, Zhetong Liang, Kwan-Yee~K Wong, and Lei Zhang.
\newblock Hdr video reconstruction: A coarse-to-fine network and a real-world benchmark dataset.
\newblock In \emph{Proceedings of the IEEE/CVF international conference on computer vision}, pages 2502--2511, 2021.

\bibitem[Chen et~al.(2022)Chen, Yang, Chan, Li, Hou, and Chau]{chen2022attention}
Jie Chen, Zaifeng Yang, Tsz~Nam Chan, Hui Li, Junhui Hou, and Lap-Pui Chau.
\newblock Attention-guided progressive neural texture fusion for high dynamic range image restoration.
\newblock \emph{IEEE Transactions on Image Processing}, 31:\penalty0 2661--2672, 2022.

\bibitem[Chen et~al.(2024)Chen, Guo, Yu, Zhang, Gu, and Xue]{chen2024event}
Yutian Chen, Shi Guo, Fangzheng Yu, Feng Zhang, Jinwei Gu, and Tianfan Xue.
\newblock Event-based motion magnification.
\newblock \emph{arXiv preprint arXiv:2402.11957}, 2024.

\bibitem[Cho et~al.(2024)Cho, Kim, Jeong, and Yoon]{cho2024tta}
Hoonhee Cho, Taewoo Kim, Yuhwan Jeong, and Kuk-Jin Yoon.
\newblock Tta-evf: Test-time adaptation for event-based video frame interpolation via reliable pixel and sample estimation.
\newblock In \emph{Proceedings of the IEEE/CVF Conference on Computer Vision and Pattern Recognition}, pages 25701--25711, 2024.

\bibitem[Chung and Cho(2023)]{chung2023lan}
Haesoo Chung and Nam~Ik Cho.
\newblock Lan-hdr: Luminance-based alignment network for high dynamic range video reconstruction.
\newblock In \emph{Proceedings of the IEEE/CVF International Conference on Computer Vision}, pages 12760--12769, 2023.

\bibitem[Debevec and Malik(2023)]{debevec2023recovering}
Paul~E Debevec and Jitendra Malik.
\newblock Recovering high dynamic range radiance maps from photographs.
\newblock In \emph{Seminal Graphics Papers: Pushing the Boundaries, Volume 2}, pages 643--652. 2023.

\bibitem[Deitke et~al.(2023)Deitke, Schwenk, Salvador, Weihs, Michel, VanderBilt, Schmidt, Ehsani, Kembhavi, and Farhadi]{deitke2023objaverse}
Matt Deitke, Dustin Schwenk, Jordi Salvador, Luca Weihs, Oscar Michel, Eli VanderBilt, Ludwig Schmidt, Kiana Ehsani, Aniruddha Kembhavi, and Ali Farhadi.
\newblock Objaverse: A universe of annotated 3d objects.
\newblock In \emph{Proceedings of the IEEE/CVF Conference on Computer Vision and Pattern Recognition}, pages 13142--13153, 2023.

\bibitem[Ding et~al.(2020)Ding, Ma, Wang, and Simoncelli]{ding2020iqa}
Keyan Ding, Kede Ma, Shiqi Wang, and Eero~P. Simoncelli.
\newblock Image quality assessment: Unifying structure and texture similarity.
\newblock \emph{CoRR}, abs/2004.07728, 2020.

\bibitem[Duan(2024)]{duan2024led}
Yuxing Duan.
\newblock Led: A large-scale real-world paired dataset for event camera denoising.
\newblock In \emph{Proceedings of the IEEE/CVF Conference on Computer Vision and Pattern Recognition}, pages 25637--25647, 2024.

\bibitem[Gardner et~al.(2017)Gardner, Sunkavalli, Yumer, Shen, Gambaretto, Gagn{\'e}, and Lalonde]{gardner2017learning}
Marc-Andr{\'e} Gardner, Kalyan Sunkavalli, Ersin Yumer, Xiaohui Shen, Emiliano Gambaretto, Christian Gagn{\'e}, and Jean-Fran{\c{c}}ois Lalonde.
\newblock Learning to predict indoor illumination from a single image.
\newblock \emph{arXiv preprint arXiv:1704.00090}, 2017.

\bibitem[Guo et~al.(2023)Guo, Chen, Gao, Yang, Bartkovjak, Qin, Hu, Zhou, Uchiyama, Kudo, et~al.]{guo20233}
Menghan Guo, Shoushun Chen, Zhe Gao, Wenlei Yang, Peter Bartkovjak, Qing Qin, Xiaoqin Hu, Dahei Zhou, Masayuki Uchiyama, Yoshiharu Kudo, et~al.
\newblock A 3-wafer-stacked hybrid 15mpixel cis+ 1 mpixel evs with 4.6 gevent/s readout, in-pixel tdc and on-chip isp and esp function.
\newblock In \emph{2023 IEEE International Solid-State Circuits Conference (ISSCC)}, pages 90--92. IEEE, 2023.

\bibitem[Guo et~al.(2019)Guo, Yan, Zhang, Zuo, and Zhang]{guo2019toward}
Shi Guo, Zifei Yan, Kai Zhang, Wangmeng Zuo, and Lei Zhang.
\newblock Toward convolutional blind denoising of real photographs.
\newblock In \emph{Proceedings of the IEEE/CVF conference on computer vision and pattern recognition}, pages 1712--1722, 2019.

\bibitem[Guo et~al.(2022)Guo, Yang, Ma, Ren, and Zhang]{guo2022differentiable}
Shi Guo, Xi Yang, Jianqi Ma, Gaofeng Ren, and Lei Zhang.
\newblock A differentiable two-stage alignment scheme for burst image reconstruction with large shift.
\newblock In \emph{Proceedings of the IEEE/CVF Conference on Computer Vision and Pattern Recognition}, pages 17472--17481, 2022.

\bibitem[Hold-Geoffroy et~al.(2019)Hold-Geoffroy, Athawale, and Lalonde]{hold2019deep}
Yannick Hold-Geoffroy, Akshaya Athawale, and Jean-Fran{\c{c}}ois Lalonde.
\newblock Deep sky modeling for single image outdoor lighting estimation.
\newblock In \emph{Proceedings of the IEEE/CVF conference on computer vision and pattern recognition}, pages 6927--6935, 2019.

\bibitem[Hu et~al.(2013)Hu, Gallo, Pulli, and Sun]{hu2013hdr}
Jun Hu, Orazio Gallo, Kari Pulli, and Xiaobai Sun.
\newblock Hdr deghosting: How to deal with saturation?
\newblock In \emph{Proceedings of the IEEE conference on computer vision and pattern recognition}, pages 1163--1170, 2013.

\bibitem[Hu et~al.(2021)Hu, Liu, and Delbruck]{hu2021v2e}
Yuhuang Hu, Shih-Chii Liu, and Tobi Delbruck.
\newblock v2e: From video frames to realistic dvs events.
\newblock In \emph{Proceedings of the IEEE/CVF Conference on Computer Vision and Pattern Recognition}, pages 1312--1321, 2021.

\bibitem[Kalantari and Ramamoorthi(2019)]{kalantari2019deep}
Nima~Khademi Kalantari and Ravi Ramamoorthi.
\newblock Deep hdr video from sequences with alternating exposures.
\newblock In \emph{Computer graphics forum}, pages 193--205. Wiley Online Library, 2019.

\bibitem[Kalantari et~al.(2013)Kalantari, Shechtman, Barnes, Darabi, Goldman, and Sen]{kalantari2013patch}
Nima~Khademi Kalantari, Eli Shechtman, Connelly Barnes, Soheil Darabi, Dan~B Goldman, and Pradeep Sen.
\newblock Patch-based high dynamic range video.
\newblock \emph{ACM Trans. Graph.}, 32\penalty0 (6):\penalty0 202--1, 2013.

\bibitem[Ke et~al.(2021)Ke, Wang, Wang, Milanfar, and Yang]{ke2021musiq}
Junjie Ke, Qifei Wang, Yilin Wang, Peyman Milanfar, and Feng Yang.
\newblock Musiq: Multi-scale image quality transformer.
\newblock In \emph{Proceedings of the IEEE/CVF international conference on computer vision}, pages 5148--5157, 2021.

\bibitem[Kim et~al.(2024)Kim, Cho, and Yoon]{kim2024frequency}
Taewoo Kim, Hoonhee Cho, and Kuk-Jin Yoon.
\newblock Frequency-aware event-based video deblurring for real-world motion blur.
\newblock In \emph{Proceedings of the IEEE/CVF Conference on Computer Vision and Pattern Recognition}, pages 24966--24976, 2024.

\bibitem[Kingma(2013)]{kingma2013auto}
Diederik~P Kingma.
\newblock Auto-encoding variational bayes.
\newblock \emph{arXiv preprint arXiv:1312.6114}, 2013.

\bibitem[Kingma and Ba(2014)]{Kingma2014AdamAM}
Diederik~P. Kingma and Jimmy Ba.
\newblock Adam: A method for stochastic optimization.
\newblock \emph{CoRR}, abs/1412.6980, 2014.

\bibitem[Kodama et~al.(2023)Kodama, Sato, Yorikado, Berner, Mizoguchi, Miyazaki, Tsukamoto, Matoba, Shinozaki, Niwa, et~al.]{kodama20231}
Kazutoshi Kodama, Yusuke Sato, Yuhi Yorikado, Raphael Berner, Kyoji Mizoguchi, Takahiro Miyazaki, Masahiro Tsukamoto, Yoshihisa Matoba, Hirotaka Shinozaki, Atsumi Niwa, et~al.
\newblock 1.22 $\mu$m 35.6 mpixel rgb hybrid event-based vision sensor with 4.88 $\mu$m-pitch event pixels and up to 10k event frame rate by adaptive control on event sparsity.
\newblock In \emph{IEEE International Solid-State Circuits Conference (ISSCC)}, pages 92--94. IEEE, 2023.

\bibitem[Kou et~al.(2017)Kou, Li, Wen, and Chen]{kou2017multi}
Fei Kou, Zhengguo Li, Changyun Wen, and Weihai Chen.
\newblock Multi-scale exposure fusion via gradient domain guided image filtering.
\newblock In \emph{2017 IEEE international conference on multimedia and expo (ICME)}, pages 1105--1110. IEEE, 2017.

\bibitem[Liu et~al.(2023)Liu, Zhang, Sun, Liang, Zeng, and Zhang]{liu2023joint}
Shuaizheng Liu, Xindong Zhang, Lingchen Sun, Zhetong Liang, Hui Zeng, and Lei Zhang.
\newblock Joint hdr denoising and fusion: A real-world mobile hdr image dataset.
\newblock In \emph{Proceedings of the IEEE/CVF Conference on Computer Vision and Pattern Recognition}, pages 13966--13975, 2023.

\bibitem[Liu et~al.(2021)Liu, Lin, Li, Rao, Jiang, Han, Fan, Sun, and Liu]{liu2021adnet}
Zhen Liu, Wenjie Lin, Xinpeng Li, Qing Rao, Ting Jiang, Mingyan Han, Haoqiang Fan, Jian Sun, and Shuaicheng Liu.
\newblock Adnet: Attention-guided deformable convolutional network for high dynamic range imaging.
\newblock In \emph{Proceedings of the IEEE/CVF Conference on Computer Vision and Pattern Recognition}, pages 463--470, 2021.

\bibitem[Liu et~al.(2022)Liu, Wang, Zeng, and Liu]{liu2022ghost}
Zhen Liu, Yinglong Wang, Bing Zeng, and Shuaicheng Liu.
\newblock Ghost-free high dynamic range imaging with context-aware transformer.
\newblock In \emph{European Conference on Computer Vision}, pages 344--360. Springer, 2022.

\bibitem[Ma and Wang(2015)]{ma2015multi}
Kede Ma and Zhou Wang.
\newblock Multi-exposure image fusion: A patch-wise approach.
\newblock In \emph{2015 IEEE International Conference on Image Processing (ICIP)}, pages 1717--1721. IEEE, 2015.

\bibitem[Ma et~al.(2017)Ma, Li, Yong, Wang, Meng, and Zhang]{ma2017robust}
Kede Ma, Hui Li, Hongwei Yong, Zhou Wang, Deyu Meng, and Lei Zhang.
\newblock Robust multi-exposure image fusion: a structural patch decomposition approach.
\newblock \emph{IEEE Transactions on Image Processing}, 26\penalty0 (5):\penalty0 2519--2532, 2017.

\bibitem[Ma et~al.(2019)Ma, Duanmu, Zhu, Fang, and Wang]{ma2019deep}
Kede Ma, Zhengfang Duanmu, Hanwei Zhu, Yuming Fang, and Zhou Wang.
\newblock Deep guided learning for fast multi-exposure image fusion.
\newblock \emph{IEEE Transactions on Image Processing}, 29:\penalty0 2808--2819, 2019.

\bibitem[Ma et~al.(2025)Ma, Guo, Chen, Xue, and Gu]{ma2025timelens}
Yongrui Ma, Shi Guo, Yutian Chen, Tianfan Xue, and Jinwei Gu.
\newblock Timelens-xl: Real-time event-based video frame interpolation with large motion.
\newblock In \emph{European Conference on Computer Vision}, pages 178--194. Springer, 2025.

\bibitem[Mangiat and Gibson(2010)]{mangiat2010high}
Stephen Mangiat and Jerry Gibson.
\newblock High dynamic range video with ghost removal.
\newblock In \emph{Applications of digital image processing XXXIII}, pages 307--314. SPIE, 2010.

\bibitem[Mertens et~al.(2007)Mertens, Kautz, and Van~Reeth]{mertens2007exposure}
Tom Mertens, Jan Kautz, and Frank Van~Reeth.
\newblock Exposure fusion.
\newblock In \emph{15th Pacific Conference on Computer Graphics and Applications (PG'07)}, pages 382--390. IEEE, 2007.

\bibitem[Messikommer et~al.(2022)Messikommer, Georgoulis, Gehrig, Tulyakov, Erbach, Bochicchio, Li, and Scaramuzza]{messikommer2022multi}
Nico Messikommer, Stamatios Georgoulis, Daniel Gehrig, Stepan Tulyakov, Julius Erbach, Alfredo Bochicchio, Yuanyou Li, and Davide Scaramuzza.
\newblock Multi-bracket high dynamic range imaging with event cameras.
\newblock In \emph{Proceedings of the IEEE/CVF conference on computer vision and pattern recognition}, pages 547--557, 2022.

\bibitem[Mildenhall et~al.(2018)Mildenhall, Barron, Chen, Sharlet, Ng, and Carroll]{mildenhall2018burst}
Ben Mildenhall, Jonathan~T Barron, Jiawen Chen, Dillon Sharlet, Ren Ng, and Robert Carroll.
\newblock Burst denoising with kernel prediction networks.
\newblock In \emph{Proceedings of the IEEE conference on computer vision and pattern recognition}, pages 2502--2510, 2018.

\bibitem[Oh et~al.(2014)Oh, Lee, Tai, and Kweon]{oh2014robust}
Tae-Hyun Oh, Joon-Young Lee, Yu-Wing Tai, and In~So Kweon.
\newblock Robust high dynamic range imaging by rank minimization.
\newblock \emph{IEEE transactions on pattern analysis and machine intelligence}, 37\penalty0 (6):\penalty0 1219--1232, 2014.

\bibitem[Pattanaik et~al.(1998)Pattanaik, Ferwerda, Fairchild, and Greenberg]{pattanaik1998multiscale}
Sumanta~N Pattanaik, James~A Ferwerda, Mark~D Fairchild, and Donald~P Greenberg.
\newblock A multiscale model of adaptation and spatial vision for realistic image display.
\newblock In \emph{Proceedings of the 25th annual conference on Computer graphics and interactive techniques}, pages 287--298, 1998.

\bibitem[Rebecq et~al.(2018)Rebecq, Gehrig, and Scaramuzza]{rebecq2018esim}
Henri Rebecq, Daniel Gehrig, and Davide Scaramuzza.
\newblock Esim: an open event camera simulator.
\newblock In \emph{Conference on robot learning}, pages 969--982. PMLR, 2018.

\bibitem[Rebecq et~al.(2019)Rebecq, Ranftl, Koltun, and Scaramuzza]{rebecq2019high}
Henri Rebecq, Ren{\'e} Ranftl, Vladlen Koltun, and Davide Scaramuzza.
\newblock High speed and high dynamic range video with an event camera.
\newblock \emph{IEEE transactions on pattern analysis and machine intelligence}, 43\penalty0 (6):\penalty0 1964--1980, 2019.

\bibitem[Rombach et~al.(2022)Rombach, Blattmann, Lorenz, Esser, and Ommer]{Rombach_2022_CVPR}
Robin Rombach, Andreas Blattmann, Dominik Lorenz, Patrick Esser, and Bj\"orn Ommer.
\newblock High-resolution image synthesis with latent diffusion models.
\newblock In \emph{Proceedings of the IEEE/CVF Conference on Computer Vision and Pattern Recognition (CVPR)}, pages 10684--10695, 2022.

\bibitem[Ronneberger et~al.(2015)Ronneberger, Fischer, and Brox]{ronneberger2015u}
Olaf Ronneberger, Philipp Fischer, and Thomas Brox.
\newblock U-net: Convolutional networks for biomedical image segmentation.
\newblock In \emph{Medical image computing and computer-assisted intervention--MICCAI 2015: 18th international conference, Munich, Germany, October 5-9, 2015, proceedings, part III 18}, pages 234--241. Springer, 2015.

\bibitem[Sen et~al.(2012)Sen, Kalantari, Yaesoubi, Darabi, Goldman, and Shechtman]{sen2012robust}
Pradeep Sen, Nima~Khademi Kalantari, Maziar Yaesoubi, Soheil Darabi, Dan~B Goldman, and Eli Shechtman.
\newblock Robust patch-based hdr reconstruction of dynamic scenes.
\newblock \emph{ACM Trans. Graph.}, 31\penalty0 (6):\penalty0 203--1, 2012.

\bibitem[Serrano-Gotarredona and Linares-Barranco(2013)]{serrano2013128}
Teresa Serrano-Gotarredona and Bernab{\'e} Linares-Barranco.
\newblock A 128×128 1.5\% contrast sensitivity 0.9\% fpn 3us latency 4 mw asynchronous frame-free dynamic vision sensor using transimpedance preamplifiers.
\newblock \emph{IEEE Journal of Solid-State Circuits}, 48\penalty0 (3):\penalty0 827--838, 2013.

\bibitem[Shaw et~al.(2022)Shaw, Catley-Chandar, Leonardis, and Perez-Pellitero]{shaw2022hdr}
Richard Shaw, Sibi Catley-Chandar, Ales Leonardis, and Eduardo Perez-Pellitero.
\newblock Hdr reconstruction from bracketed exposures and events.
\newblock \emph{arXiv preprint arXiv:2203.14825}, 2022.

\bibitem[Su et~al.(2020)Su, Yan, Zhu, Zhang, Ge, Sun, and Zhang]{su2020blindly}
Shaolin Su, Qingsen Yan, Yu Zhu, Cheng Zhang, Xin Ge, Jinqiu Sun, and Yanning Zhang.
\newblock Blindly assess image quality in the wild guided by a self-adaptive hyper network.
\newblock In \emph{Proceedings of the IEEE/CVF conference on computer vision and pattern recognition}, pages 3667--3676, 2020.

\bibitem[Teed and Deng(2020)]{teed2020raft}
Zachary Teed and Jia Deng.
\newblock Raft: Recurrent all-pairs field transforms for optical flow.
\newblock In \emph{Computer Vision--ECCV 2020: 16th European Conference, Glasgow, UK, August 23--28, 2020, Proceedings, Part II 16}, pages 402--419. Springer, 2020.

\bibitem[Tulyakov et~al.(2021)Tulyakov, Gehrig, Georgoulis, Erbach, Gehrig, Li, and Scaramuzza]{tulyakov2021time}
Stepan Tulyakov, Daniel Gehrig, Stamatios Georgoulis, Julius Erbach, Mathias Gehrig, Yuanyou Li, and Davide Scaramuzza.
\newblock Time lens: Event-based video frame interpolation.
\newblock In \emph{Proceedings of the IEEE/CVF conference on computer vision and pattern recognition}, pages 16155--16164, 2021.

\bibitem[Wang et~al.(2019{\natexlab{a}})Wang, Ho, Yoon, et~al.]{wang2019event}
Lin Wang, Yo-Sung Ho, Kuk-Jin Yoon, et~al.
\newblock Event-based high dynamic range image and very high frame rate video generation using conditional generative adversarial networks.
\newblock In \emph{Proceedings of the IEEE/CVF Conference on Computer Vision and Pattern Recognition}, pages 10081--10090, 2019{\natexlab{a}}.

\bibitem[Wang et~al.(2019{\natexlab{b}})Wang, Chan, Yu, Dong, and Change~Loy]{wang2019edvr}
Xintao Wang, Kelvin~CK Chan, Ke Yu, Chao Dong, and Chen Change~Loy.
\newblock Edvr: Video restoration with enhanced deformable convolutional networks.
\newblock In \emph{Proceedings of the IEEE/CVF conference on computer vision and pattern recognition workshops}, pages 0--0, 2019{\natexlab{b}}.

\bibitem[Wang et~al.(2024)Wang, Wang, Gong, and Xue]{wang2024bilateral}
Yuehao Wang, Chaoyi Wang, Bingchen Gong, and Tianfan Xue.
\newblock Bilateral guided radiance field processing.
\newblock \emph{ACM Transactions on Graphics (TOG)}, 43\penalty0 (4):\penalty0 1--13, 2024.

\bibitem[Wu et~al.(2018)Wu, Xu, Tai, and Tang]{wu2018deep}
Shangzhe Wu, Jiarui Xu, Yu-Wing Tai, and Chi-Keung Tang.
\newblock Deep high dynamic range imaging with large foreground motions.
\newblock In \emph{Proceedings of the European Conference on Computer Vision (ECCV)}, pages 117--132, 2018.

\bibitem[Xiao et~al.(2002)Xiao, DiCarlo, Catrysse, and Wandell]{xiao2002high}
Feng Xiao, Jeffrey~M DiCarlo, Peter~B Catrysse, and Brian~A Wandell.
\newblock High dynamic range imaging of natural scenes.
\newblock In \emph{Color and imaging conference}, pages 337--342. Society of Imaging Science and Technology, 2002.

\bibitem[Xiaopeng et~al.(2024)Xiaopeng, Zhaoyuan, Cien, Chen, Lei, and Lei]{xiaopeng2024hdr}
Li Xiaopeng, Zeng Zhaoyuan, Fan Cien, Zhao Chen, Deng Lei, and Yu Lei.
\newblock Hdr imaging for dynamic scenes with events.
\newblock \emph{arXiv preprint arXiv:2404.03210}, 2024.

\bibitem[Xu et~al.(2024)Xu, Wang, Gu, Xue, and Yang]{xu2024hdrflow}
Gangwei Xu, Yujin Wang, Jinwei Gu, Tianfan Xue, and Xin Yang.
\newblock Hdrflow: Real-time hdr video reconstruction with large motions.
\newblock In \emph{CVPR}, 2024.

\bibitem[Yan et~al.(2019)Yan, Gong, Shi, Hengel, Shen, Reid, and Zhang]{yan2019attention}
Qingsen Yan, Dong Gong, Qinfeng Shi, Anton van~den Hengel, Chunhua Shen, Ian Reid, and Yanning Zhang.
\newblock Attention-guided network for ghost-free high dynamic range imaging.
\newblock In \emph{Proceedings of the IEEE/CVF Conference on Computer Vision and Pattern Recognition}, pages 1751--1760, 2019.

\bibitem[Yan et~al.(2022)Yan, Gong, Shi, Van Den~Hengel, Shen, Reid, and Zhang]{yan2022dual}
Qingsen Yan, Dong Gong, Javen~Qinfeng Shi, Anton Van Den~Hengel, Chunhua Shen, Ian Reid, and Yanning Zhang.
\newblock Dual-attention-guided network for ghost-free high dynamic range imaging.
\newblock \emph{International Journal of Computer Vision}, pages 1--19, 2022.

\bibitem[Yan et~al.(2023)Yan, Chen, Zhang, Zhu, Sun, and Zhang]{yan2023unified}
Qingsen Yan, Weiye Chen, Song Zhang, Yu Zhu, Jinqiu Sun, and Yanning Zhang.
\newblock A unified hdr imaging method with pixel and patch level.
\newblock In \emph{Proceedings of the IEEE/CVF Conference on Computer Vision and Pattern Recognition}, pages 22211--22220, 2023.

\bibitem[Yang et~al.(2022)Yang, Wu, Shi, Lao, Gong, Cao, Wang, and Yang]{yang2022maniqa}
Sidi Yang, Tianhe Wu, Shuwei Shi, Shanshan Lao, Yuan Gong, Mingdeng Cao, Jiahao Wang, and Yujiu Yang.
\newblock Maniqa: Multi-dimension attention network for no-reference image quality assessment.
\newblock In \emph{Proceedings of the IEEE/CVF Conference on Computer Vision and Pattern Recognition}, pages 1191--1200, 2022.

\bibitem[Yang et~al.(2023)Yang, Han, Liang, Sato, and Shi]{yang2023learning}
Yixin Yang, Jin Han, Jinxiu Liang, Imari Sato, and Boxin Shi.
\newblock Learning event guided high dynamic range video reconstruction.
\newblock In \emph{Proceedings of the IEEE/CVF Conference on Computer Vision and Pattern Recognition}, pages 13924--13934, 2023.

\bibitem[Zhang et~al.(2023)Zhang, Rao, and Agrawala]{zhang2023adding}
Lvmin Zhang, Anyi Rao, and Maneesh Agrawala.
\newblock Adding conditional control to text-to-image diffusion models.
\newblock In \emph{Proceedings of the IEEE/CVF International Conference on Computer Vision}, pages 3836--3847, 2023.

\bibitem[Zhang et~al.(2018)Zhang, Isola, Efros, Shechtman, and Wang]{zhang2018perceptual}
Richard Zhang, Phillip Isola, Alexei~A Efros, Eli Shechtman, and Oliver Wang.
\newblock The unreasonable effectiveness of deep features as a perceptual metric.
\newblock In \emph{CVPR}, 2018.

\bibitem[Zhang et~al.(2024)Zhang, Ma, Chen, Zhang, Gu, Xue, and Guo]{zhang2024sim}
Ziran Zhang, Yongrui Ma, Yueting Chen, Feng Zhang, Jinwei Gu, Tianfan Xue, and Shi Guo.
\newblock From sim-to-real: Toward general event-based low-light frame interpolation with per-scene optimization.
\newblock \emph{arXiv preprint arXiv:2406.08090}, 2024.

\end{thebibliography}
}

\clearpage
\setcounter{page}{1}
\maketitlesupplementary

\appendix
\renewcommand\thefigure{A\arabic{figure}}
\renewcommand\thetable{A\arabic{table}}  
\renewcommand\theequation{A\arabic{equation}}
\setcounter{section}{0}
\setcounter{equation}{0}
\setcounter{table}{0}
\setcounter{figure}{0}

\section{More details and discussion}
\subsection{Drawback of $\mu$-law tone mapping}
Although the $\mu$-law tone mapping is effective for standard HDR images, it proves insufficient for handling both bright and dark regions in 12-stop HDR images, as demonstrated in \cref{fig:supp_1}. Using only $\mu$-law tone-mapped HDR as the loss function is inadequate for accurately recovering fine details in HDR images. To illustrate this, we present a toy example in \cref{fig:supp_1} (b), where a small Gaussian noise with $\sigma=2\times10^{-5}$ is added to the HDR image. While the noisy HDR image appears nearly identical to the clean HDR under $\mu$-law tone mapping, a more advanced local tone mapping approach, such as the "Vibrant" style in commercial software, reveals significant noise-induced texture degradation.

Furthermore, compared to Event-HDR* (\cref{fig:supp_1} (c)), which uses training data nearly the same with ours, the HDR results under $\mu$-law tone mapping may appear slightly smoother but do not exhibit noticeably more artifacts than our method. However, under the "Vibrant" tone mapping style, Event-HDR* introduces visible artifacts, underscoring the limitations of relying solely on $\mu$-law tone mapping as a loss function for 12-stop HDR reconstruction.

To address this issue, we incorporate a pre-trained diffusion model, leveraging its priors from natural images. As shown in \cref{fig:supp_1} (d) and (e), our method achieves sharper detail recovery while maintaining artifact-free results even after local tone mapping.

\subsection{Color correction for HSHDR dataset}
\begin{figure}[h!]
    \centering
\includegraphics[width=0.47\textwidth]{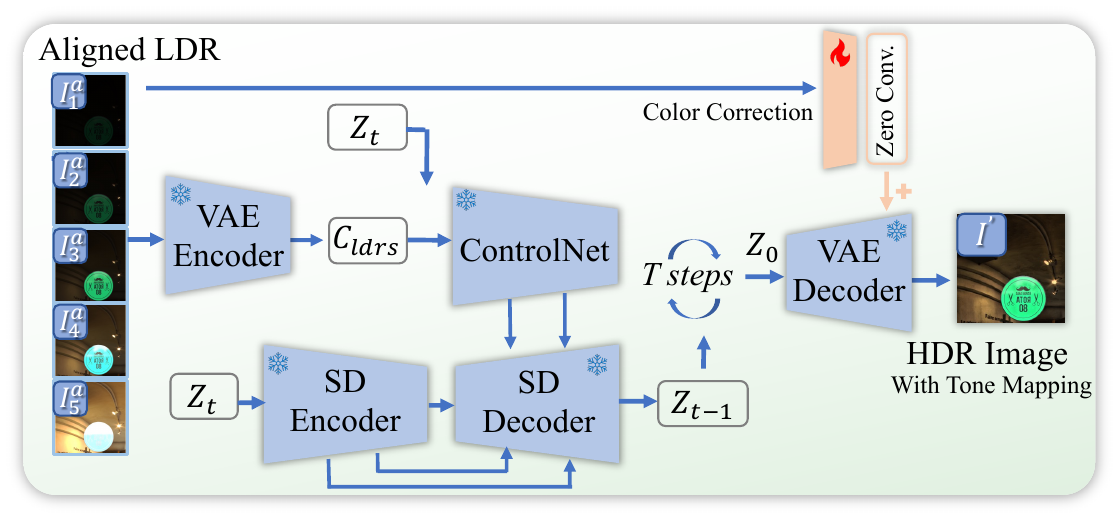}
    \vspace{-0pt}
    \caption{Color correction module for diffusion-based fusion module.}
    \label{fig:color-collection-framework}
\end{figure}

\begin{figure*}[h!]
    \centering
\includegraphics[width=1.0\textwidth]{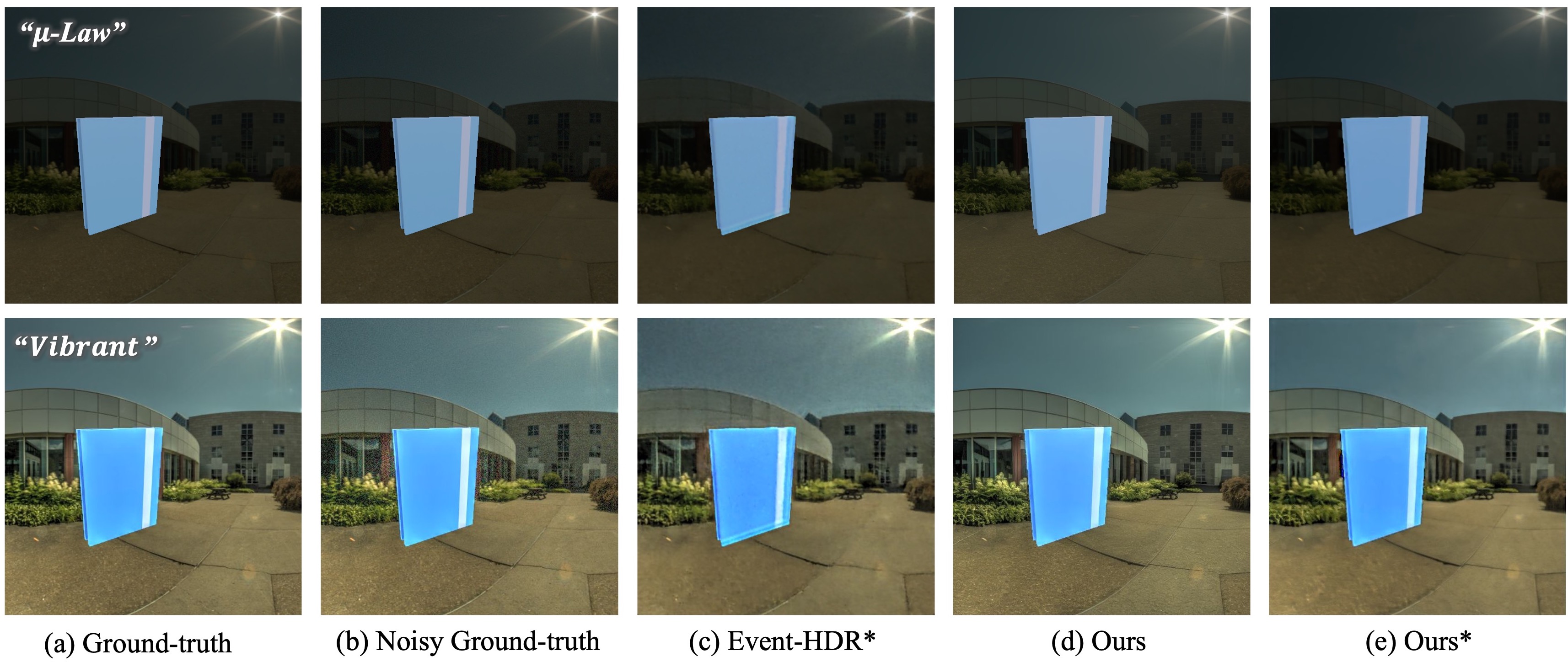}
    \vspace{-10pt}
    \caption{Comparison of results under the commonly used $\mu$-law tone mapping style and the \emph{Vibrant} tone mapping style from commercial HDR software (Photomatix). (b) represents the noisy ground-truth image, obtained by adding Gaussian noise with a very small noise level ($2\times10^{-5}$) to (a). \textbf{Please zoom in for more details.}}
    \label{fig:supp_1}
\end{figure*}

From \cref{fig:supp_1} (a), it is evident that the ground truth with $\mu$-law tone mapping for 12-stop HDR appears excessively dark, mainly because the high-luminance regions (\eg, the sun). This discrepancy might arise from a mismatch between the $\mu$-law tone mapping and the priors learned by the diffusion model. We observed that the diffusion-based fusion produces HDR outputs with $\mu$-law tone mapping that are a little brighter than ground-truth, leading to decreased performance on metrics such as PSNR on the HSHDR dataset. To address this, we adopted a strategy of training one layer of the VAE decoder for color correction, denoted as Ours*. The framework is illustrated in \cref{fig:color-collection-framework}. 

The input to the color correction module comprises the aligned LDR images. The output features are processed through a "zero-initialized convolutional" layer, ensuring that the training's initial state does not interfere with the outputs of the preceding diffusion-based fusion module. These features are then integrated into the original VAE decoder. Importantly, only the color correction module is updated during training, while the parameters of the VAE, ControlNet, and diffusion modules are kept frozen. 

As shown in \cref{fig:supp_1}, our approach (Ours*) effectively corrects the color shifting introduced by the diffusion process in the $\mu$-law tone mapping. It is also worth noting that in local tone mapping styles (\eg, "\emph{Vibrant}"), the original model (\cref{fig:supp_1} (d)) does not exhibit noticeable color shifting compared to the ground truth. Our original HDR results are capable of accurately recovering the relative brightness and contrast between pixels. By comparing \cref{fig:supp_1} (d) and (e), we observe that while updating one layer of the VAE decoder enhances brightness accuracy, it also slightly compromises the priors learned by the diffusion model, resulting in marginally blurrier HDR textures in Ours* compared to Ours. Nevertheless, compared to the previous state-of-the-art method (\ie, Event-HDR*), our approach demonstrates superior performance, particularly in reducing ghosting artifacts and preserving fine details.

\subsection{Real-world alignment finetuning}
\begin{figure}[h!]
    \centering
\includegraphics[width=0.47\textwidth]{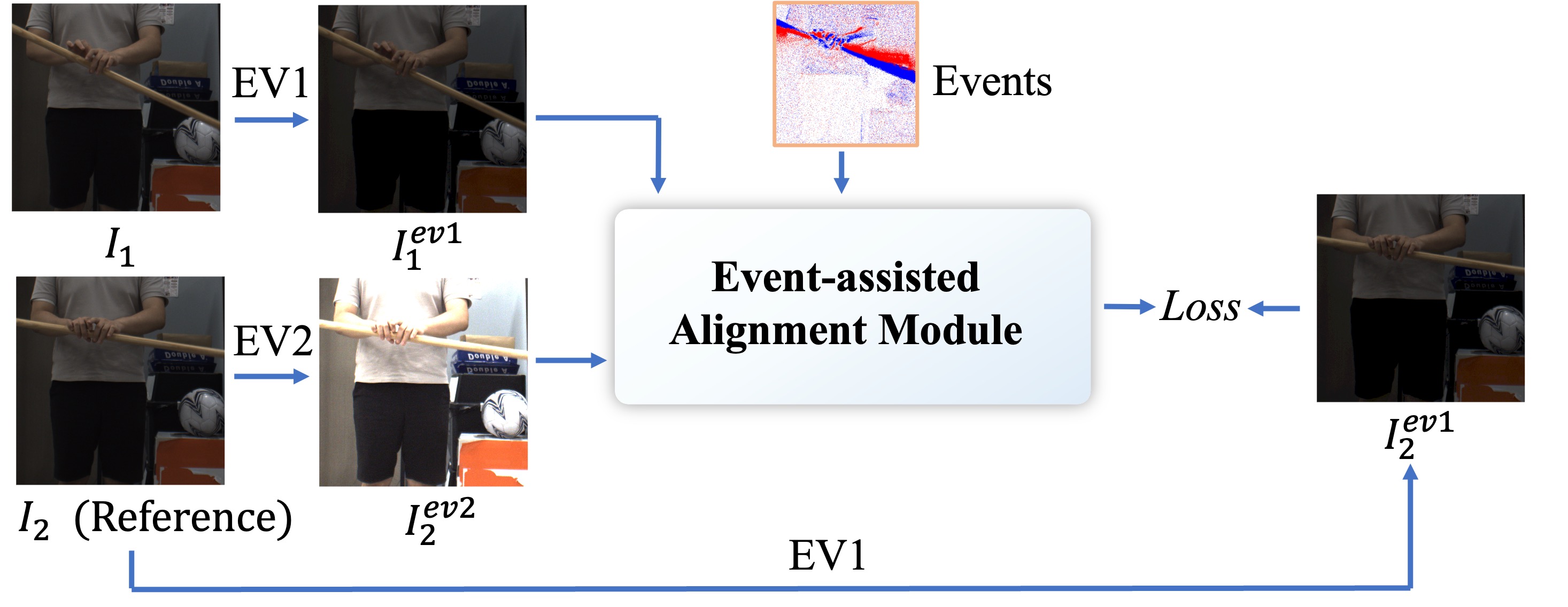}
    \vspace{-0pt}
    \caption{Event-assisted alignment module fine-tuning.}
    \label{fig:finetune-framework}
\end{figure}

In this section, we provide additional details on using a real-world interpolation dataset to fine-tune the event-assisted alignment module. The framework is illustrated in \cref{fig:finetune-framework}. The two input images, $I_1$ and $I_2$, are first converted into $I_1^{ev1}$ and $I_2^{ev2}$, respectively. Due to the exposure gap, $I_1^{ev1}$ and $I_2^{ev2}$ provide incomplete and inconsistent image information, with certain regions in $I_1^{ev1}$ being underexposed and some areas in $I_2^{ev2}$ being overexposed. This enables the network to leverage real event signals to learn alignment information effectively.

As shown in \cref{fig:ablation} in the main text, fine-tuning with real-world interpolation data improves the alignment performance on real event datasets. This refinement further reduces ghosting artifacts in the subsequent fusion results.

\subsection{Event-RGB camera system}

\begin{figure}[h!]
    \centering
\includegraphics[width=0.30\textwidth]{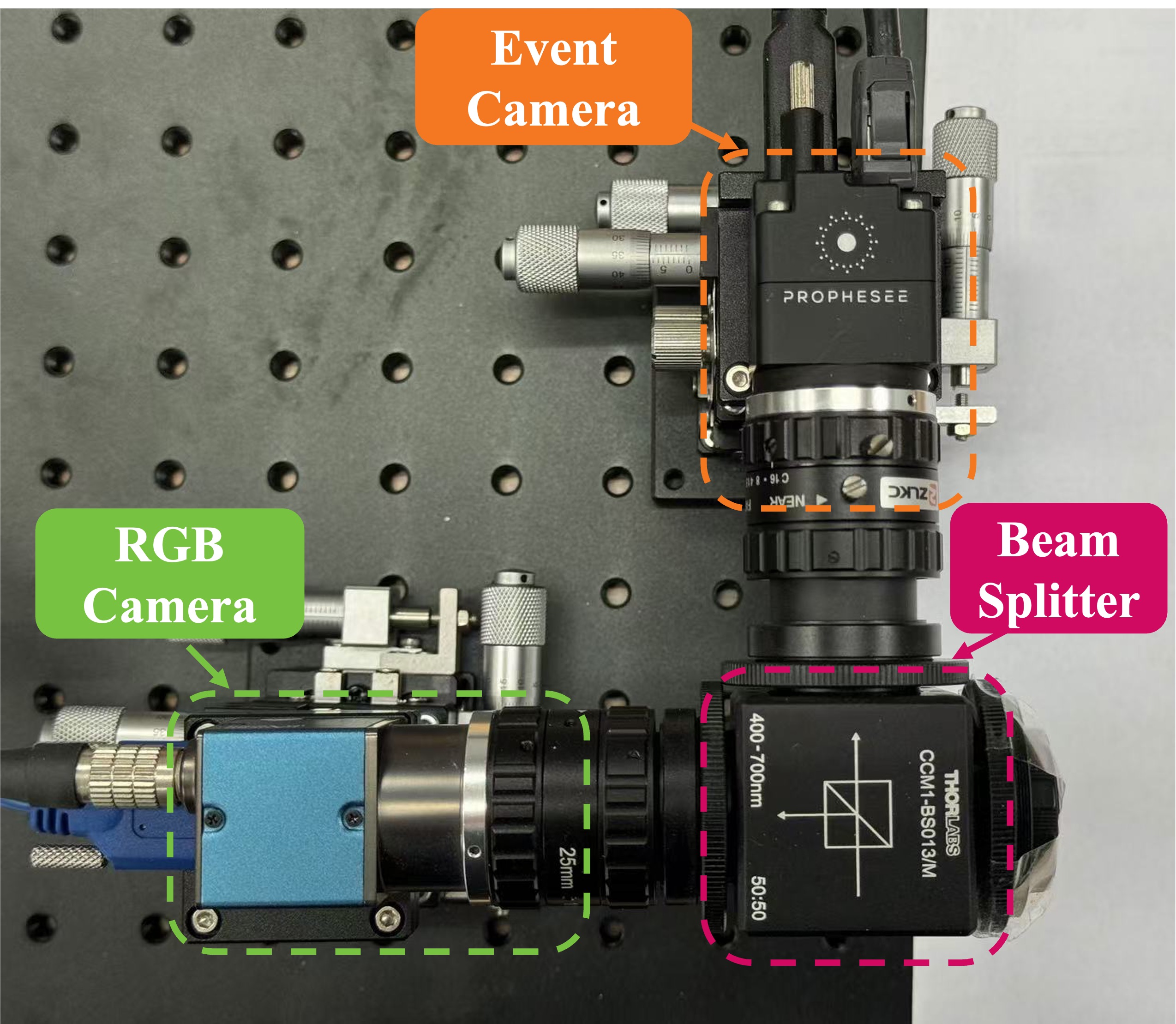}
    \vspace{-0pt}
    \caption{Our Event-RGB dual camera system.}
    \label{fig:camera-system}
\end{figure}

Our Event-RGB dual-camera system, illustrated in \cref{fig:camera-system}, utilizes a beam splitter to align the fields of view of the event and RGB cameras. Two cameras are time-synchronized using a hardware trigger. To achieve spatial alignment between event and RGB images, a flickering checkerboard is used for accurate calibration.

%\begin{figure}[h!]
%\setlength{\abovecaptionskip}{-0.0cm}
%\setlength{\belowcaptionskip}{-0.2cm}
%    \centering
%\includegraphics[width=0.5\textwidth]%{CVPR25_EVSUDR/images/dis_hdr_trans.jpg}
%    \vspace{-0pt}
%    \caption{HDR-Transformer HDR results by using different EV images.}
%    \label{fig:dis-hdr-trans}
%\end{figure}

%\subsection{Limitations of previous deep dynamic HDR}

%In this section, we provide a simple discussion on why previous dynamic HDR methods have notiable ghosting artifacts. In \cref{fig:dis-hdr-trans}, we provide the HDR results of HDR-Transformer by using [] 

\section{More real-world visual comparisons}
\begin{figure*}[t!]
\setlength{\belowcaptionskip}{-0.0cm}
    \centering
\includegraphics[width=\textwidth]{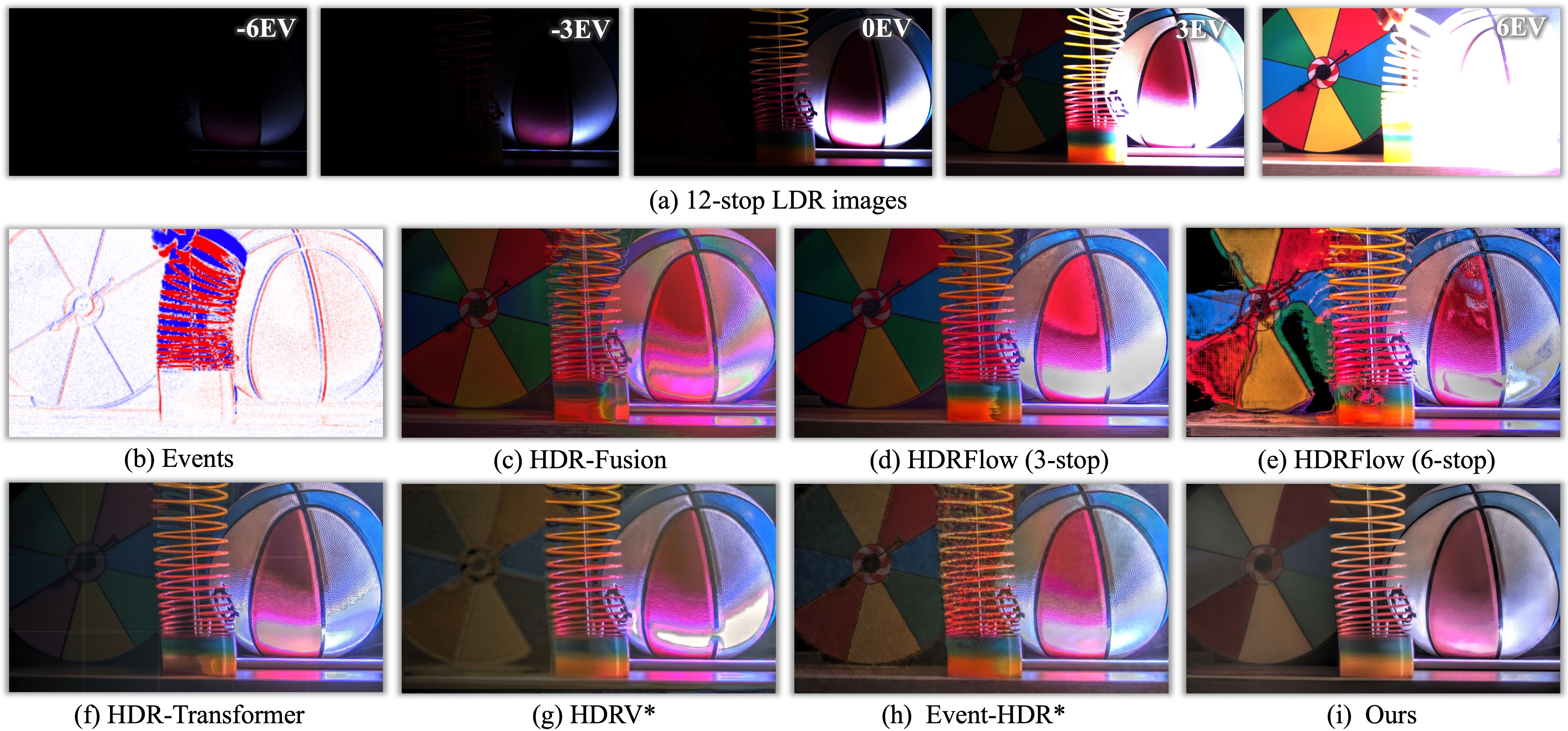}
    \vspace{-20pt}
    \caption{Visual comparisons of different HDR imaging methods on real-captured data. We utilize tone mapping in commercial HDR software (Photomatix) to better visualization.}
    \label{fig:exp_supp_real_1}
\end{figure*}

\begin{figure*}[t!]
\setlength{\belowcaptionskip}{-0.0cm}
    \centering
\includegraphics[width=\textwidth]{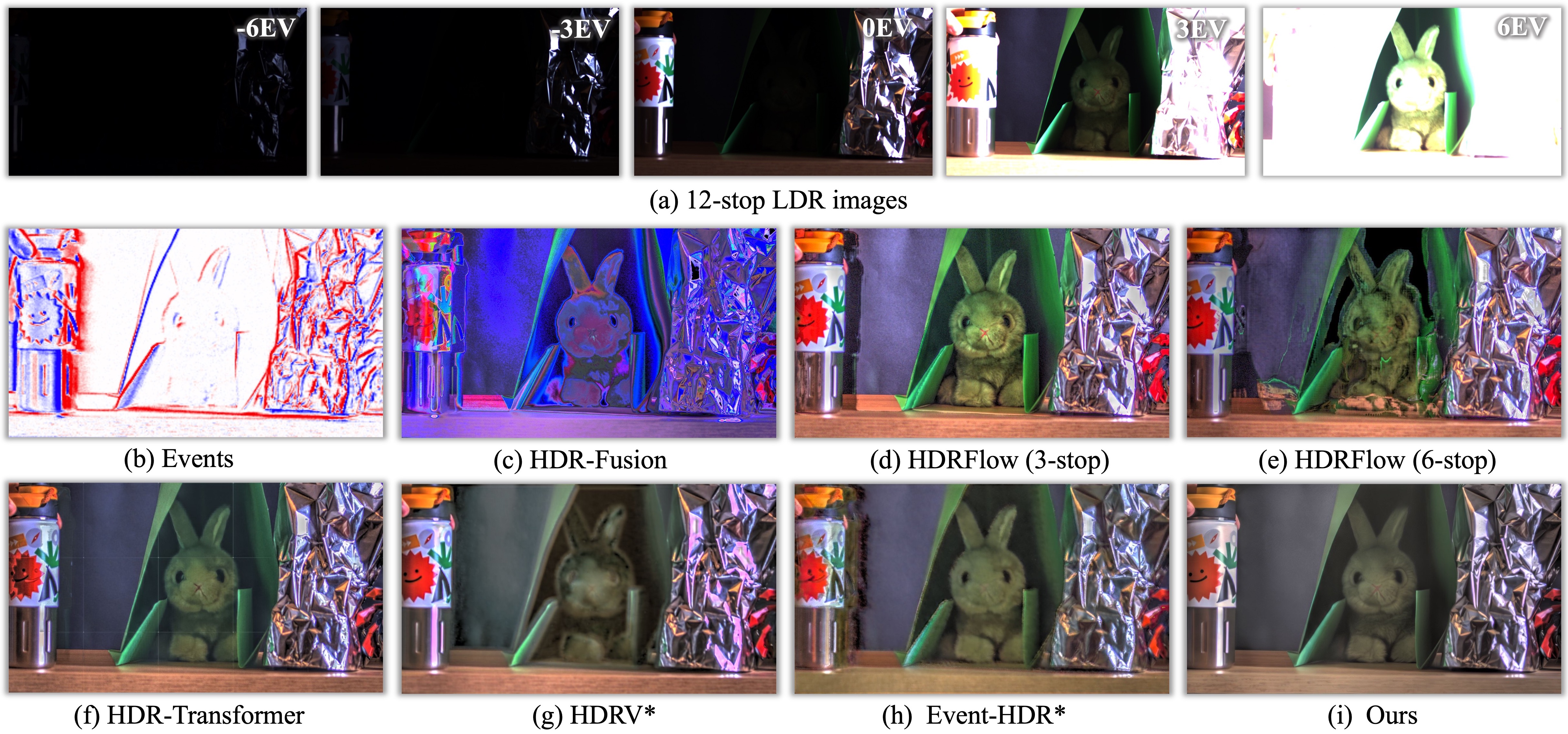}
    \vspace{-20pt}
    \caption{Visual comparisons of different HDR imaging methods on real-captured data. We utilize tone mapping in commercial HDR software (Photomatix) to better visualization.}
    \label{fig:exp_supp_real_2}
\end{figure*}

More visual comparisons of real-world experiments are presented in \cref{fig:exp_supp_real_1,fig:exp_supp_real_2}. As shown, due to the lack of motion information between LDR images, previous dynamic HDR methods, such as HDRFlow and HDR-Transformer, struggle to align LDRs with significant exposure differences effectively, leading to ghosting artifacts in the resulting HDR images. For the HDRV* method, although using event signals to enhance single-frame LDR images helps avoid ghosting, the absence of color information in event signals, coupled with the temporal accumulation of event data, results in issues such as inaccurate color reconstruction and edge bleeding. Similarly, while Event-HDR* utilizes event signals, the performance is limited in regions with large motion due to the gap between real-world and simulated event data. In contrast, our method, with its explicitly designed alignment module and fine-tuning on real interpolated datasets, effectively minimizes HDR ghosting artifacts. Additionally, benefiting from the powerful image priors of the diffusion model, our fusion method achieves natural 12-stop HDR results, even in regions with large exposure differences.

\end{document}